\documentclass[11pt]{article}

\usepackage[a4paper,left=0.92in,right=0.92in,top=0.86in,bottom=0.86in,headheight=14pt,headsep=10pt,footskip=20pt]{geometry}
\usepackage[T1]{fontenc}
\usepackage[utf8]{inputenc}
\usepackage{tgpagella}
\usepackage{microtype}
\usepackage[hyphens]{url}
\usepackage{graphicx}
\usepackage{booktabs}
\usepackage{multirow}
\usepackage{amsmath,amssymb}
\usepackage{xcolor}
\usepackage{framed}
\usepackage{float}
\usepackage{caption}
\usepackage{natbib}
\usepackage{xspace}
\usepackage{enumitem}
\usepackage{hyperref}

\makeatletter
\renewcommand\section{\@startsection{section}{1}{\z@}%
  {-3.2ex\@plus -0.8ex\@minus -0.2ex}%
  {1.8ex\@plus 0.3ex}%
  {\normalfont\Large\bfseries}}
\renewcommand\subsection{\@startsection{subsection}{2}{\z@}%
  {-2.4ex\@plus -0.6ex\@minus -0.2ex}%
  {1.1ex\@plus 0.2ex}%
  {\normalfont\large\bfseries}}
\renewcommand\subsubsection{\@startsection{subsubsection}{3}{\z@}%
  {-1.8ex\@plus -0.4ex\@minus -0.1ex}%
  {0.7ex\@plus 0.1ex}%
  {\normalfont\normalsize\bfseries}}
\def\ps@giftbench{%
  \def\@oddhead{\vbox{\hrule height 0.35pt\vskip 3pt\hbox to\textwidth{\hfil\small\itshape GIFTBench and ForenScope\hfil}}}%
  \def\@evenhead{\@oddhead}%
  \def\@oddfoot{\hfil\small\thepage\hfil}%
  \let\@evenfoot\@oddfoot}
\makeatother
\definecolor{abstractbg}{RGB}{248,242,237}
\definecolor{abstractborder}{RGB}{226,198,183}
\colorlet{shadecolor}{abstractbg}
\renewenvironment{abstract}{%
  \begin{center}\begin{minipage}{0.94\linewidth}
  \begin{shaded}
  \small
  \setlength{\parindent}{0pt}
}{%
  \end{shaded}\end{minipage}\end{center}
}

\hypersetup{
  colorlinks=true,
  linkcolor=black,
  citecolor=black,
  urlcolor=blue,
  hypertexnames=false,
  pdftitle={GIFTBench: Diagnosing Generalization in Image Forgery Localization and Informing Model Design},
  pdfauthor={Baoke Dou, Ziye Wang, Hao Wang, Guoqing Cai, Wende Tan, Chenyang Si, Liucheng Guo, Yueming Lyu}
}

\newcommand{\gift}{\textbf{GIFTBench}\xspace}
\newcommand{\method}{\textsc{ForenScope}\xspace}

\title{GIFTBench: Diagnosing Generalization in Image Forgery Localization and Informing Model Design}
\author{Baoke Dou, Ziye Wang, Hao Wang, Guoqing Cai,\\
Wende Tan, Chenyang Si, Liucheng Guo, Yueming Lyu}
\date{}

\begin{document}
\maketitle

\begin{abstract}
Reliable evaluation of image forgery localization (IFL) requires assessing models under diverse distribution changes, yet existing benchmarks often cover limited manipulation conditions or entangle multiple factors in cross-dataset evaluation. Consequently, aggregate performance provides an incomplete view of localization generalization. We introduce \gift, a multi-axis benchmark of 115,013 manipulated images with pixel-level annotations spanning manipulation source, semantic target, editing operation, and composition complexity. GIFTBench supports axis-specific transfer analysis and evaluation on twelve external datasets. Its diagnostic studies reveal asymmetric cross-source transfer, recall-dominated failures, and heterogeneous degradation across semantic, operational, and compositional changes. Beyond diagnosis, the scale and diversity of GIFTBench provide a substantially broader training distribution than conventional IFL datasets. Training representative localizers on GIFTBench consistently improves their aggregate transfer to external datasets, showing that the benchmark serves not only as an evaluation tool but also as an effective training resource for cross-domain localization. Guided by the diagnostic findings, we further develop \method, a detection and localization framework combining classification-adapted representations with multi-depth, multi-scale spatial features, learned layer fusion, and selective coarse-scale conditioning. Experiments show improved cross-dataset localization while retaining image-level detection capability. The GIFTBench dataset showcase page is available at \url{https://giftbench-preview.doudoudouya337.chatgpt.site}.
\end{abstract}

\section{Introduction}

Image manipulation tools have made it increasingly easy to alter image content while preserving a visually plausible appearance. A complete forensic system should therefore answer two complementary questions: \emph{whether} an image is real or manipulated and \emph{where} the manipulation occurs. The former corresponds to image-level manipulation detection, while the latter is typically formulated as image forgery localization (IFL), which predicts a pixel-level manipulation mask. Although recent methods achieve strong performance on established benchmarks, existing evaluation does not fully characterize how they generalize under diverse distribution changes beyond their training data. That distribution shift reduces performance is unsurprising; the more informative question is \emph{which change makes a trained localizer fail}.

Existing datasets vary in image content, manipulation source, semantic target, editing operation, and composition \citep{dong2013casia,hsu2006detecting,wen2016coverage,guan2019mfc,novozamsky2020imd2020,tralic2013comofod}, but these factors are commonly entangled across datasets or only partially annotated within them. As Table~\ref{tab:benchmark-comparison} shows, many benchmarks already contain large collections of classical or AI-based edits and substantial manipulation diversity. The limitation is therefore not simply dataset scale, but how this diversity is represented and used. Without a shared factorized annotation space, cross-dataset evaluation can reveal that performance drops but provides limited evidence about which variation is associated with the failure. At the same time, a useful benchmark should provide sufficiently diverse and consistently annotated data to serve as a reusable training resource rather than only an analysis tool. We therefore argue for a benchmark that jointly represents multiple manipulation factors, supports axis-specific evaluation, and provides a broad training distribution for studying and improving cross-domain localization.

\begin{table}[t]
\centering
\fontsize{4.6pt}{5.0pt}\selectfont
\setlength{\tabcolsep}{1.8pt}
\renewcommand{\arraystretch}{0.86}
\caption{Comparison of representative image forgery localization benchmarks. \(\checkmark\) denotes an explicitly represented factor and \(\circ\) a factor present but not separately evaluated. GIFTBench scale reports manipulated images; other scales follow the original benchmarks.}

\label{tab:benchmark-comparison}
\resizebox{0.94\linewidth}{!}{%
\begin{tabular}{lcccccc}
\toprule
Benchmark & Scale & Classical & AI-based & Semantic Target & Operation & Composition \\
\midrule
CASIA v2~\citep{dong2013casia} & 12.6K & \checkmark & $\times$ & $\circ$ & $\circ$ & $\times$ \\
DEFACTO~\citep{mahfoudi2019defacto} & 159K & \checkmark & $\times$ & $\circ$ & $\circ$ & $\times$ \\
IMD2020~\citep{novozamsky2020imd2020} & 40K & \checkmark & $\times$ & $\circ$ & $\circ$ & $\times$ \\
CocoGlide~\citep{guillaro2023trufor} & 20K & $\times$ & \checkmark & $\circ$ & $\circ$ & $\times$ \\
AutoSplice~\citep{jia2023autosplice} & 110K & $\times$ & \checkmark & $\circ$ & $\circ$ & $\times$ \\
TGIF~\citep{mareen2024tgif} & 60K & $\times$ & \checkmark & $\circ$ & $\circ$ & $\circ$ \\
GIM~\citep{chen2025gim} & 2.28M & $\times$ & \checkmark & $\circ$ & $\circ$ & $\times$ \\
BR-Gen~\citep{cai2026zooming} & 150K & $\times$ & \checkmark & \checkmark & $\circ$ & $\times$ \\
DiffSeg30k~\citep{ci2025diffseg30k} & 30K & $\times$ & \checkmark & $\circ$ & \checkmark & \checkmark \\
ManipBench~\citep{xu2025manipshield} & 450K+ & $\times$ & \checkmark & $\circ$ & $\circ$ & $\times$ \\
LocateEdit-Bench~\citep{wu2026locateedit} & 231K & $\times$ & \checkmark & $\circ$ & \checkmark & $\times$ \\
OpenSDI~\citep{wang2025opensdi} & 300K & $\times$ & \checkmark & $\circ$ & $\circ$ & $\times$ \\
\midrule
\textbf{GIFTBench} & \textbf{115,013} & \textbf{\checkmark} & \textbf{\checkmark} & \textbf{\checkmark} & \textbf{\checkmark} & \textbf{\checkmark} \\
\bottomrule
\end{tabular}%
}
\end{table}

To support both structured evaluation and diverse model training, we introduce \gift, a multi-axis benchmark built from ADE20K with 115,013 manipulated images and pixel-level masks. Each sample is annotated by manipulation source, semantic target, editing operation, and composition complexity, covering diverse scenes together with both classical and AI-based manipulations. Rather than constructing separate datasets for individual factors, GIFTBench organizes these variations within a unified generation and annotation framework, yielding a coherent training distribution with diversity across multiple levels. The structured annotations enable axis-specific transfer analysis, while the complete training set can be used directly to train localization models. Generalization beyond the benchmark is further evaluated on twelve external datasets.

The axis-specific protocols reveal asymmetric cross-source transfer, recall-dominated failures, and heterogeneous degradation across semantic, operational, and compositional shifts. Beyond diagnosis, training representative localizers on the complete GIFTBench set improves aggregate transfer to external datasets over their reference training distributions, showing that its diversity benefits both generalization evaluation and transferable model learning.

We develop \method, a joint detection--localization framework built on a classification-adapted DINOv2 representation \citep{oquab2024dinov2}. Its localization branch combines multi-depth patch features, learnable cross-layer fusion, bounded global-to-local modulation, and coarse-to-fine decoding; the design rationale and empirical tests are presented later.

Our contributions are threefold:
\begin{itemize}[leftmargin=*, topsep=2pt, itemsep=1pt, parsep=0pt, partopsep=0pt]

    \item We introduce \textbf{\gift}, a large-scale, multi-axis benchmark with diverse scenes and manipulations organized within a unified construction framework, supporting both structured generalization evaluation and model training.

    \item We show that GIFTBench enables systematic diagnosis of localization generalization, revealing asymmetric cross-source transfer, recall-dominated failures, and heterogeneous semantic, operational, and compositional degradation. Training on the complete GIFTBench distribution also improves aggregate transfer across external datasets.

    \item We develop and evaluate \textbf{\method}, a detection and localization model that retains image-level detection while improving cross-dataset localization.

\end{itemize}

\section{Related Work}

\paragraph{Image manipulation localization benchmarks.}
Early IFL benchmarks focused on splicing, copy-move, and limited post-processing artifacts \citep{dong2013casia,hsu2006detecting,wen2016coverage,guan2019mfc}. Later datasets broadened manipulation sources and included more realistic in-the-wild images \citep{novozamsky2020imd2020,tralic2013comofod}. Subsequent benchmarks further increased scale and manipulation diversity \citep{mahfoudi2019defacto,kadam2021multiple}, while recent work has increasingly addressed generative and instruction-driven editing \citep{jia2023autosplice,mareen2024tgif,chen2025gim,cai2026zooming}.

\paragraph{IFL models.}
IFL systems use constrained convolutions, noise or frequency cues, patch reasoning, transformer encoders, and segmentation decoders \citep{bayar2018constrained,wu2019mantra,chen2021image,hu2020span,wang2022objectformer,ma2023iml}. More recent approaches use foundation-model representations and stronger multimodal or generative priors \citep{guillaro2023trufor,su2025can,zhu2025mesoscopic,ma2024imdl}.

\paragraph{Foundation models for dense prediction.}

Large self-supervised vision encoders provide semantically rich features for dense prediction. General segmentation systems have shown that transformer representations can support accurate masks when combined with task-specific decoders \citep{kirillov2023segment,ravi2024sam}. In image forensics, however, semantically invariant representations may suppress weak manipulation-specific evidence.

\section{GIFTBench: Multi-Axis Benchmark for Generalization Diagnosis}
\label{sec:benchmark}

\subsection{Benchmark Design}

GIFTBench studies image forgery localization under benchmark-defined distribution shifts by annotating manipulation variation along four explicit dimensions and evaluating generalization separately with respect to each. Its design follows three principles: major sources of shift should be explicitly labeled rather than hidden within a random split; manipulation masks should have deterministic provenance; and the benchmark should support both axis-specific analysis and transfer to established external datasets.

Accordingly, each manipulated image is characterized along four complementary axes:
\begin{itemize}[leftmargin=*, topsep=2pt, itemsep=0pt, parsep=0pt, partopsep=0pt]
    \item \textbf{Manipulation source:} splicing, copy--move, removal, and AI editing;
    \item \textbf{Semantic target:} object-, component-, and background-level manipulation;
    \item \textbf{Editing operation:} adding, removing, and replacing content;
    \item \textbf{Composition complexity:} single manipulation, fusion of multiple edits, and multi-source composition.
\end{itemize}
Manipulation source describes the editing pipeline or generation family used to produce a sample, whereas editing operation describes the intended transformation applied to the image. The two annotations are distinct: \emph{Removal} denotes a source-family pipeline, while \emph{Remove} denotes an editing objective that may also be realized by other source families. Each sample retains both labels and can therefore be evaluated from either diagnostic perspective.
The first three axes characterize the origin, spatial semantics, and intent of an edit, while the composition axis captures the interaction of multiple manipulation processes or regions within the same image. Together, these factors define the benchmark's multi-axis evaluation space.

\begin{figure}[t]
\centering
\includegraphics[width=\linewidth]{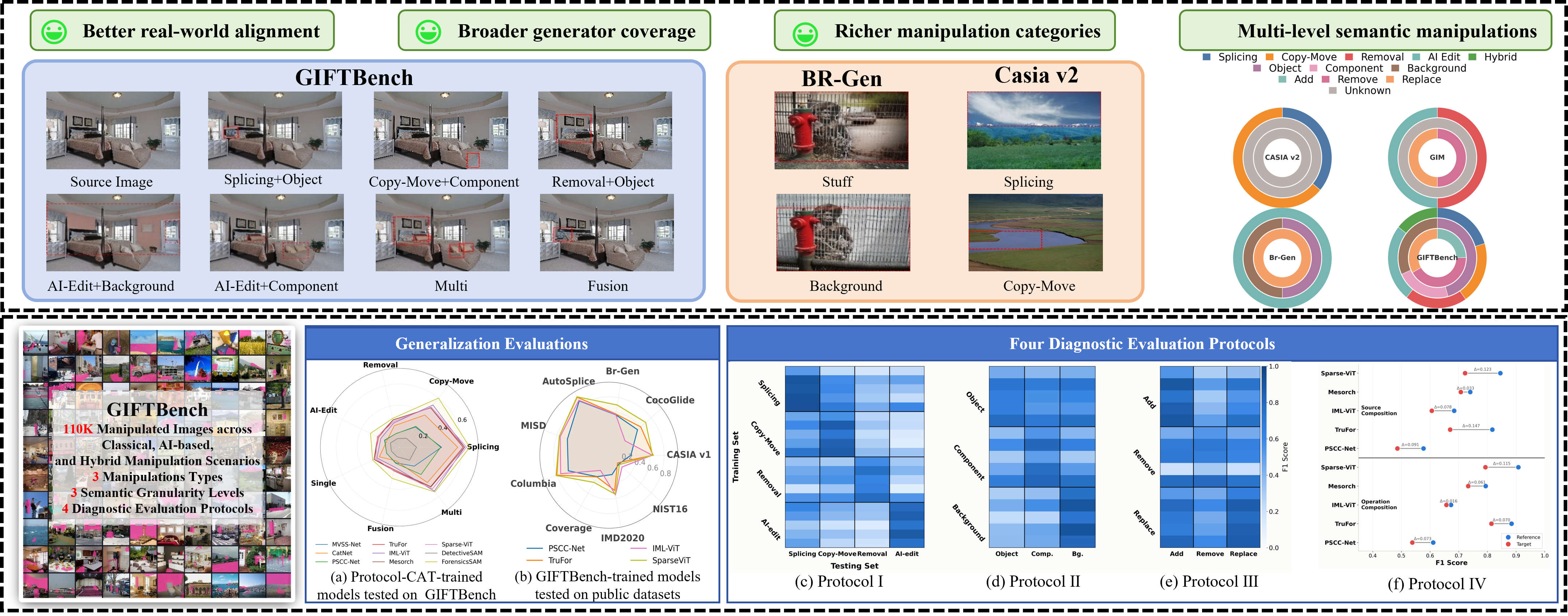}
\caption{Overview of GIFTBench, summarizing benchmark positioning, data construction, and diagnostic analysis across four generalization axes.}
\label{fig:gift_overview}
\end{figure}

\begin{figure}[t]
    \centering
    \includegraphics[width=0.96\linewidth]{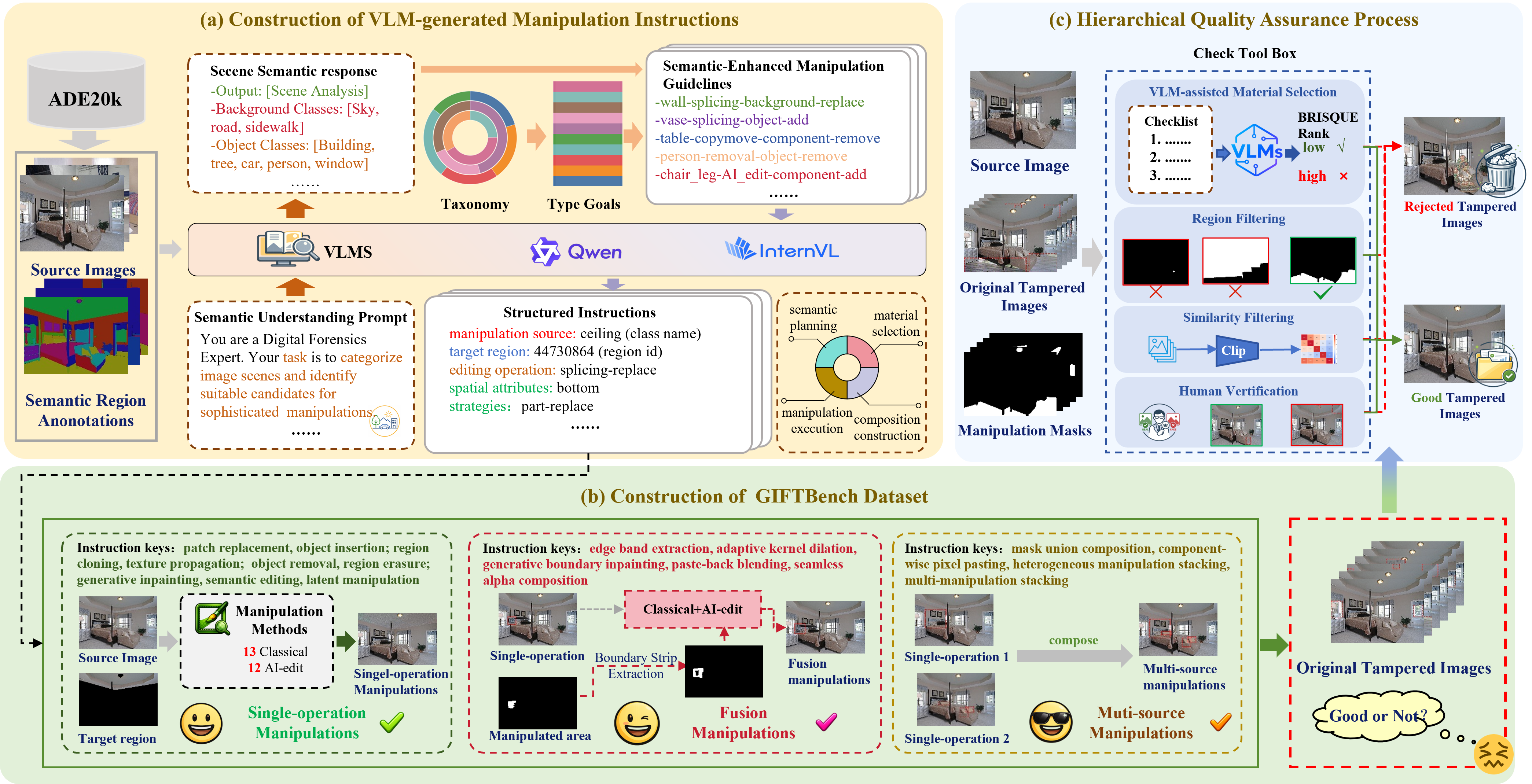}
    \caption{Construction pipeline of GIFTBench.}
    \label{fig:pipeline}
\end{figure}

\paragraph{Construction summary.}
GIFTBench uses ADE20K images and semantic annotations~\citep{zhou2019semantic} to generate manipulations through vision--language-guided planning and multiple editing pipelines, while retaining explicit masks and provenance verified by automatic and manual quality checks. Before generation, all primary and donor sources are assigned to disjoint train or test pools, which are inherited by every generated sample and its source material. The benchmark contains 115,013 manipulated images, including 103,319 training and 11,694 test images; Fig.~\ref{fig:pipeline} summarizes the construction process, with full quality-control and release details provided in Appendix~\ref{app:benchmark}.

\subsection{Evaluation Protocols}

GIFTBench defines four axis-specific diagnostic protocols. Each organizes training and evaluation along one benchmark axis under common evaluation rules, without requiring the remaining annotations to be statistically matched. Results should therefore be interpreted as axis-associated transfer patterns rather than strict one-variable-at-a-time interventions.

\begin{itemize}[leftmargin=*, topsep=2pt, itemsep=1pt, parsep=0pt, partopsep=0pt]

    \item \textbf{Protocol I: Manipulation-source transfer.}
    Models are trained on one source and evaluated on four manipulation sources, yielding a \(4\times4\) transfer matrix.

    \item \textbf{Protocol II: Semantic Target transfer.}
    Models are trained and evaluated across three semantic target levels, yielding a \(3\times3\) transfer matrix.

    \item \textbf{Protocol III: Editing-operation transfer.}
    Models are trained and tested across three operations, yielding a \(3\times3\) transfer matrix.

    \item \textbf{Protocol IV: Composition generalization.}
    Models are trained on single-operation examples and evaluated on single, fusion, and multi-source manipulations to assess generalization to composed settings.

\end{itemize}
In addition to the axis-specific protocols, we evaluate two forms of external transfer: released checkpoints are applied directly to GIFTBench without adaptation, while GIFTBench-trained models are evaluated on external datasets without target-domain fine-tuning. The former diagnoses the OOD transfer of existing systems, whereas the latter assesses generalization beyond the GIFTBench construction domain.
Pixel-level F1 and IoU are averaged over manipulated images, and cross-dataset results are macro-averaged over datasets. Real images are excluded from localization metrics; image-level detection is evaluated separately on datasets containing both real and manipulated images.

\subsection{Benchmark Diagnosis}
\label{sec:diagnosis}

\paragraph{Existing models exhibit a substantial transfer gap.}
We analyze axis-specific failures by retraining representative CNN-, Transformer-, and foundation-model-based localizers with identical source-image splits, thresholds, and metrics. For released-checkpoint transfer, we use Protocol-MVSS and Protocol-CAT checkpoints from IMDL-BenCo~\citep{ma2024imdl} and the MMTD-Set checkpoint from FakeShield~\citep{xu2024fakeshield}. Transfer is markedly stronger on Splicing and Copy--Move than on Removal and AI--Edit. The same architecture behaves differently across training protocols, while model rankings vary among source and composition groups. Thus, OOD performance depends on both the training distribution and target condition and cannot be adequately characterized by a single aggregate score. Complete results are provided in Appendix Table~\ref{tab:giftbench-public-transfer}.

\begin{figure}[t]
    \centering
    \begin{minipage}[t]{0.92\textwidth}
        \centering
    \begin{minipage}[t]{0.98\linewidth}
        \centering
        \includegraphics[width=\linewidth]{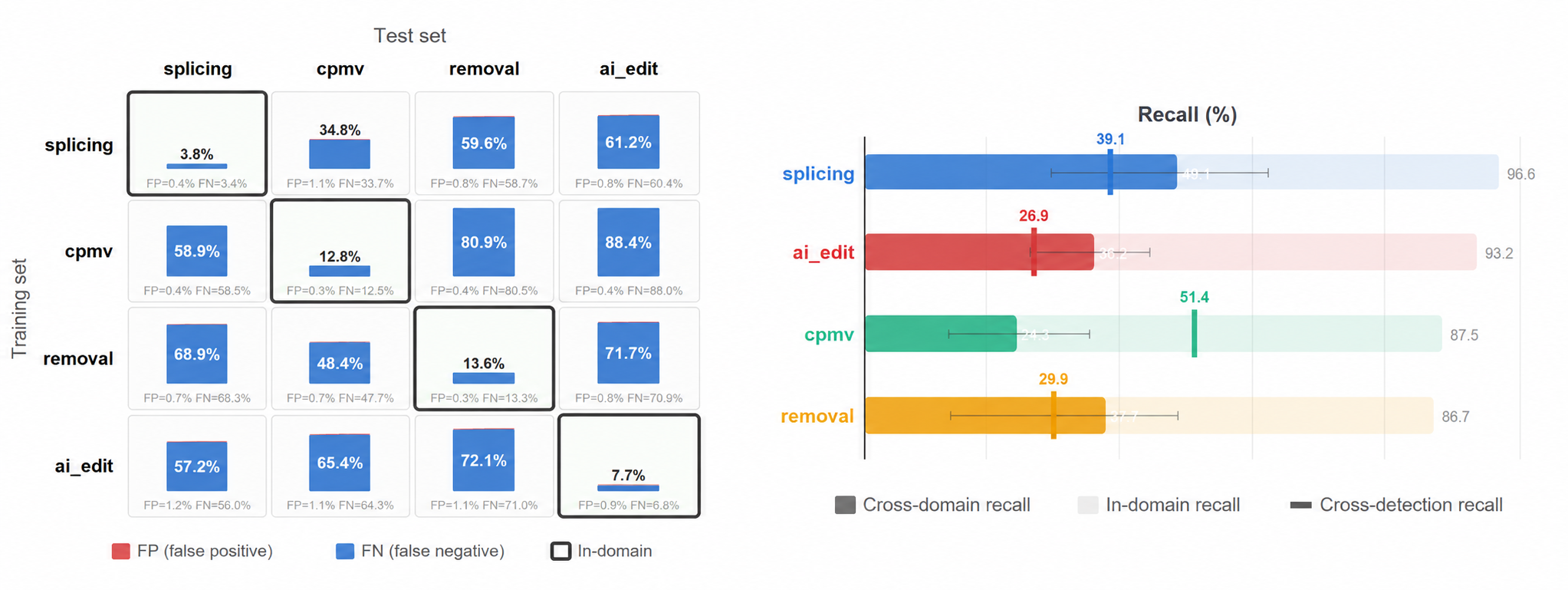}
        \par
        {\footnotesize (a) Source error and recall decomposition\par}
    \end{minipage}

    \begin{minipage}[t]{0.34\textwidth}
        \centering
        \includegraphics[width=\linewidth]{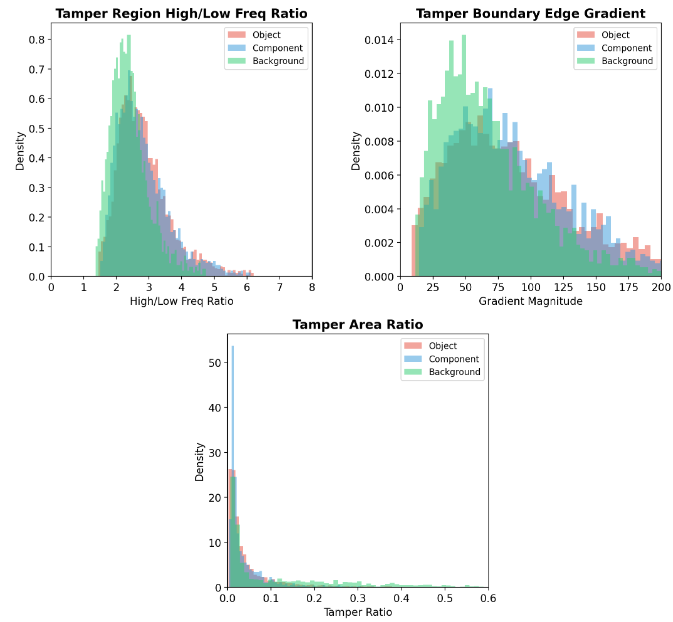}
        \par
        {\scriptsize (b) Semantic-target region statistics\par}
    \end{minipage}\hfill
    \begin{minipage}[t]{0.31\textwidth}
        \centering
        \includegraphics[width=\linewidth]{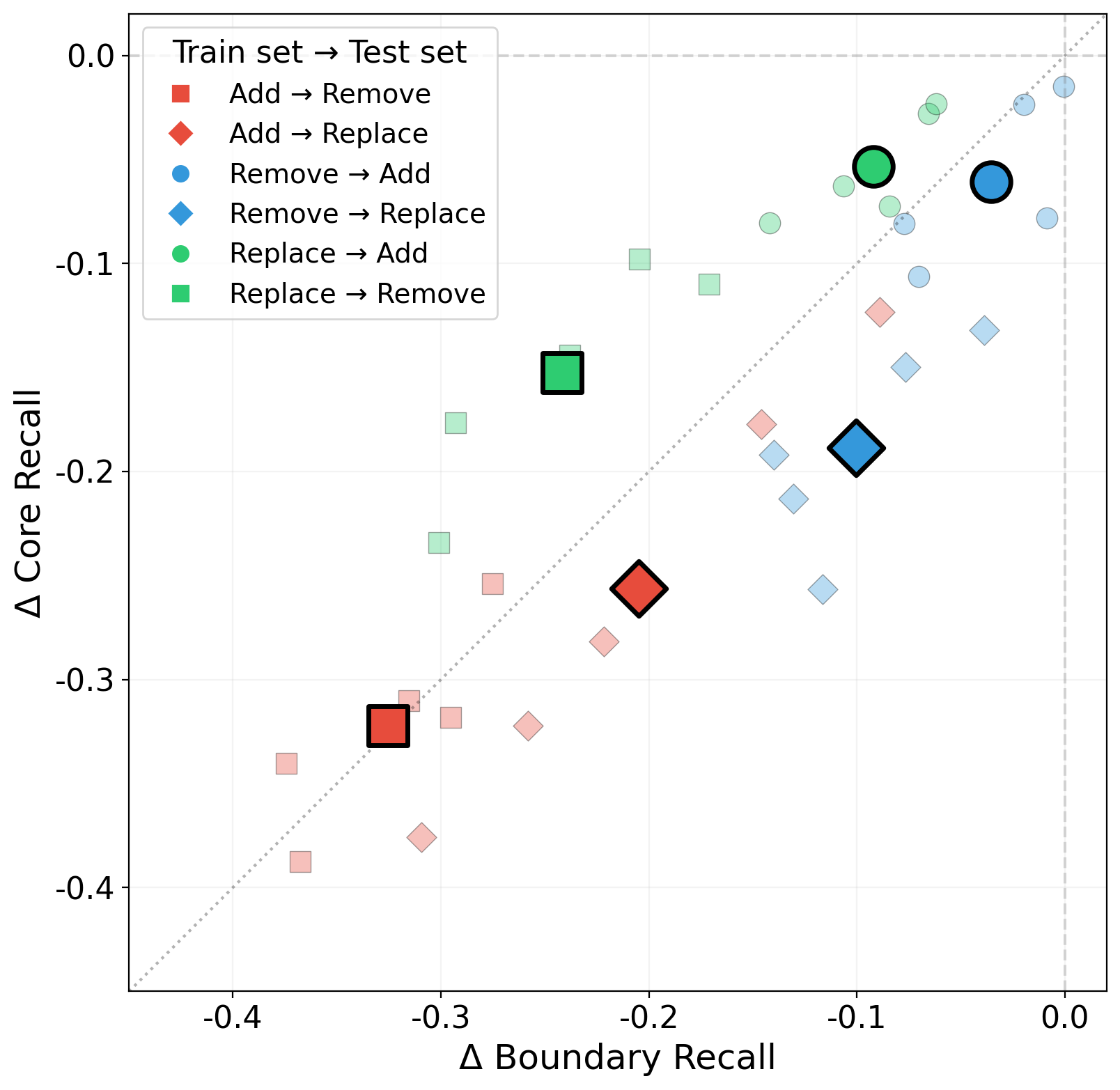}
        \par
        {\scriptsize (c) Boundary/core recall degradation\par}
    \end{minipage}\hfill
    \begin{minipage}[t]{0.31\textwidth}
        \centering
        \includegraphics[width=\linewidth]{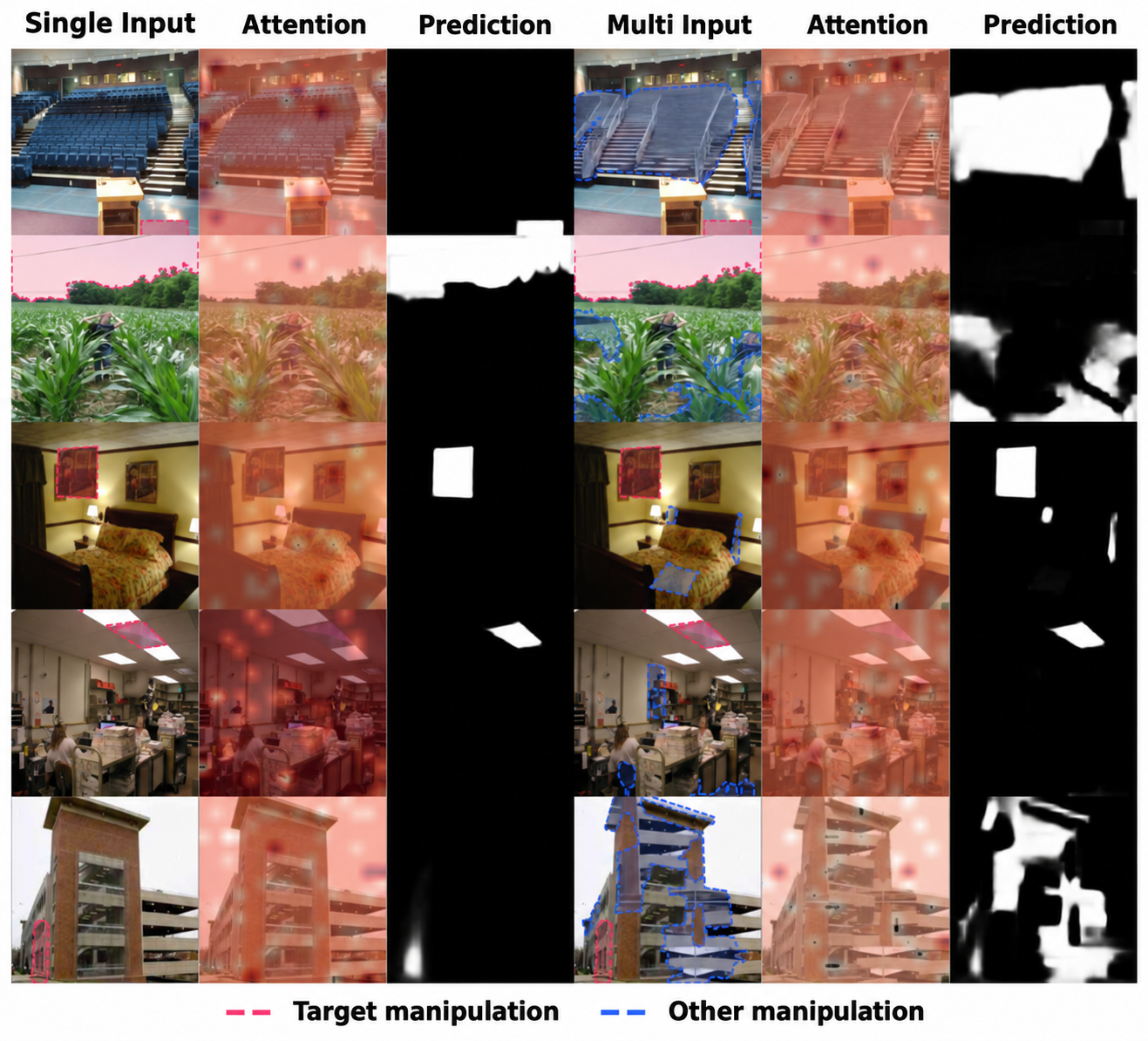}
        \par
        {\scriptsize (d) Attention under composition\par}
    \end{minipage}
    \end{minipage}
    \caption{Failure analyses for source, semantic-target, operation, and composition shifts in GIFTBench.}
    \label{fig:benchmark-diagnosis-main}
\end{figure}

Across the source, semantic-target, and operation protocols, transfer varies with the train--test pairing; Appendix Fig.~\ref{fig:diagnosis-details} provides a compact visualization of these cross-domain asymmetries.
Figure~\ref{fig:benchmark-diagnosis-main} analyzes the failure mechanisms beyond aggregate scores. In panel (a), off-diagonal source transfers are dominated by false negatives while false positives remain low, and recall drops sharply relative to matched evaluation. Unfamiliar source traces are therefore chiefly missed rather than producing indiscriminate foreground predictions. Panel (b) shows that semantic-target groups also differ in their forensic statistics: background edits tend to occupy larger regions and exhibit weaker boundary gradients and lower high-frequency energy than object- and component-level edits. The semantic transfer gap thus coincides with changes in spatial extent, texture, and boundary strength rather than with category labels alone.

Panel (c) further reveals operation-dependent spatial failures. Add-trained models show relatively balanced boundary and core recall degradation, Remove-trained models lose more core recall, and Replace-trained models lose more boundary recall, indicating different reliance on interior and boundary evidence. Controlled single-to-composition evaluation also degrades across models (Appendix Table~\ref{tab:benchmark_composition}); panel (d) illustrates how co-occurring manipulations can redirect SparseViT attention from the original target toward competing visual regions, producing incomplete or expanded predictions. Together, these analyses show that generalization fails through distinct spatial mechanisms: missed unfamiliar cues, altered region statistics, boundary/interior sensitivity, and attention competition under composition.

\subsection{External Generalization of GIFTBench-trained Models}
\label{sec:benchmark-external}

To assess generalization beyond GIFTBench, we evaluate representative localizers trained under GIFTBench and two established protocols on twelve external datasets. Table~\ref{tab:benchmark-external} reports pixel-level F1 across nine conventional IFL datasets and three related manipulation-localization datasets. GIFTBench training improves the aggregate external performance of all evaluated architectures relative to their corresponding reference training distributions. In particular, SparseViT trained on GIFTBench achieves the strongest Avg.-All F1 of 0.6029 among the evaluated benchmark-trained baselines. The gains are nevertheless target dependent, indicating that no single training distribution uniformly dominates across all external datasets.

Protocol-CAT includes IMD2020 during training; its IMD2020 result is therefore excluded from external evaluation. These results complement the axis-specific diagnosis above: GIFTBench is not only useful for diagnosing generalization failures, but also provides a diverse training distribution that transfers effectively to established external benchmarks.

\begin{table}[t]
\centering
\scriptsize
\setlength{\tabcolsep}{3pt}
\renewcommand{\arraystretch}{0.9}
\caption{Cross-dataset localization performance under different training protocols.}
\label{tab:benchmark-external}
\resizebox{\linewidth}{!}{%
\begin{tabular}{llccccccccc ccc cc}
\toprule
\multirow{2}{*}{Training Protocol} & \multirow{2}{*}{Model}
& \multicolumn{9}{c}{IFL Dataset} & \multicolumn{3}{c}{Related Dataset} & \multicolumn{2}{c}{Average} \\
\cmidrule(lr){3-11}\cmidrule(lr){12-14}\cmidrule(lr){15-16}
&& CASIA v1 & NIST16 & IMD2020 & Coverage & Columbia & MISD & AutoSplice & CocoGlide & Br-Gen & DDL & T-IC13 & OSTF & Avg. (All) & Avg. (Excl. IMD) \\
\midrule
\multirow{3}{*}{\textbf{Protocol-MVSS}}
& PSCC-Net & .375&.184&.259&.239&.605&.690&.659&.285&.267 & .358&.254&.190 & .3638&-- \\
& TruFor & .721&.321&.323&.424&.865&.756&.320&.204&.085 & .486&.364&.230 & .4249&-- \\
& IML-ViT & .718&.291&.322&.438&.747&.710&.221&.211&.079 & .312&.370&.239 & .3882&-- \\
\midrule
\multirow{4}{*}{\textbf{Protocol-CAT}}
& PSCC-Net & .569&.344&--&.386&.856&.763&.564&.509&.133 & .348&.173&.240 & --&.4441 \\
& TruFor & .821&.301&--&.483&.876&.772&.327&.283&.078 & .596&.259&.218 & --&.4558 \\
& IML-ViT & .801&.488&--&.659&.945&.710&.252&.315&.096 & .475&.415&.286 & --&.4947 \\
& SparseViT & .820&.324&--&.543&.957&.767&.385&.387&.096 & .570&.291&.214 & --&.4867 \\
\midrule
\multirow{4}{*}{\textbf{GIFTBench}}
& PSCC-Net & .320&.189&.271&.339&.649&.641&.856&.567&.563 & .415&.461&.285 & .4630{\scriptsize(+27\%)}&.4805{\scriptsize(+8\%)} \\
& TruFor & .642&.164&.490&.344&.801&.763&.917&.603&.574 & .533&.527&.368 & .5605{\scriptsize(+32\%)}&.5669{\scriptsize(+24\%)} \\
& IML-ViT & .628&.251&.552&.241&.742&.749&.868&.306&.568 & .468&.481&.316 & .5145{\scriptsize(+33\%)}&.5107{\scriptsize(+3\%)} \\
& SparseViT & .633&.200&.543&.430&.813&.763&.923&.698&.643 & .710&.521&.358 & .6029&.6084{\scriptsize(+25\%)} \\
\bottomrule
\end{tabular}%
}
\end{table}

\section{ForenScope: Benchmark-Informed Detection and Localization}
\label{sec:method}

\subsection{Overview}

The GIFTBench diagnosis further indicates that manipulation evidence can be source sensitive, spatially heterogeneous under composition, and easily missed when local cues become ambiguous. These observations do not uniquely determine an architecture; instead, they suggest three design requirements: avoiding dependence on a single representation, modeling evidence across spatial extents, and allowing uncertain local predictions to exploit complementary image-level context. ForenScope instantiates these requirements as a joint detection--localization model with multi-depth layer--scale fusion, bounded CLS conditioning, and coarse-to-fine decoding.

\begin{figure}[t]
\centering
\IfFileExists{figures/method_overview.png}{%
\includegraphics[width=0.92\linewidth]{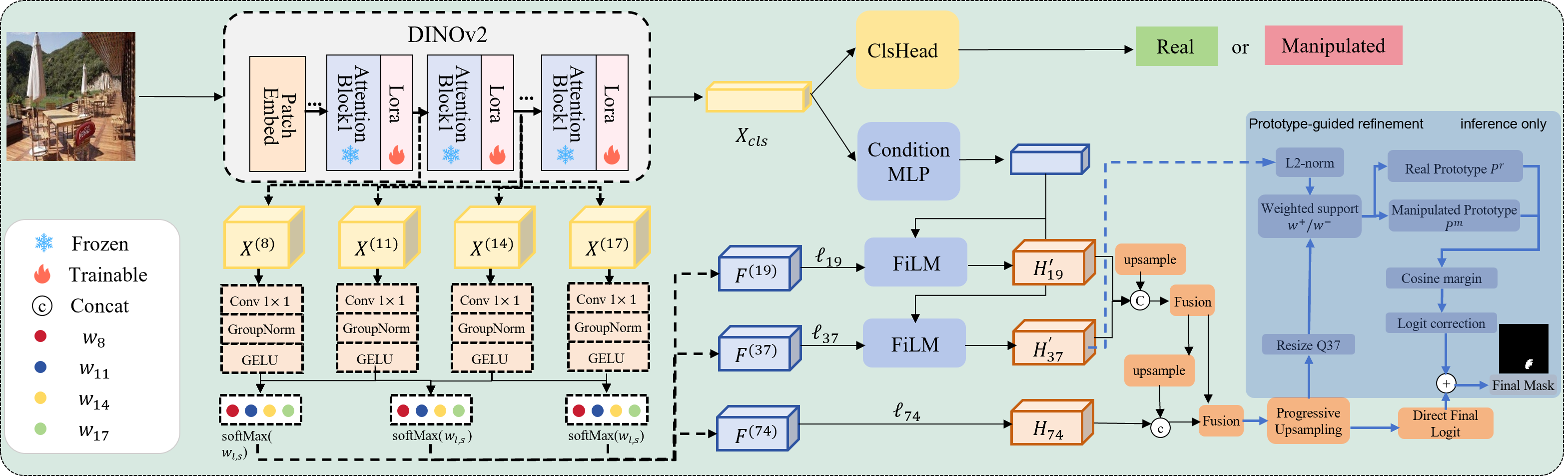}%
}{%
\fbox{\parbox{0.92\linewidth}{\centering
Method overview figure to be inserted.}}
}
\caption{Overview of ForenScope. The model jointly performs image-level detection and pixel-level localization through a frozen classification-adapted encoder and a multi-scale localization decoder.}
\label{fig:method}
\end{figure}

As illustrated in Fig.~\ref{fig:method}, ForenScope builds on a classification-adapted DINOv2 encoder. Its CLS representation supports image-level detection, while intermediate patch representations retain spatial evidence for localization. The two tasks remain separately predicted, that allows the model to exploit task complementarity without making localization depend on a hard image-level decision.

The inherited classification head predicts the image-level manipulation probability as
\begin{equation}
p_{\mathrm{det}}(I)
=
\sigma\!\left(
\mathrm{ClsHead}(z_{\mathrm{cls}})
\right).
\end{equation}
For localization, patch features from four encoder depths are projected to three spatial scales and fused by independent scale-wise LayerMixers. The class representation modulates the coarser features through bounded residual Feature-wise Linear Modulation (FiLM), after which a coarse-to-fine decoder produces the direct manipulation mask. This learned network output is the default ForenScope prediction; an optional prototype-guided refinement is studied separately at inference time.

\subsection{Multi-Depth Layer--Scale Fusion}

Manipulation cues need not be represented most strongly at the same encoder depth or spatial resolution. We therefore extract spatial patch features \(X^{(l)}\) from four depths, \(\mathcal{L}=\{8,11,14,17\}\), and project every depth to three localization scales, \(\mathcal{S}=\{74,37,19\}\). For each depth--scale pair, an independent lateral node maps the feature from 1024 to 128 channels and to the target resolution using a \(1\times1\) convolution, GroupNorm, GELU, and scale-specific resizing, producing \(U_{l,s}\).

At each scale, an independent LayerMixer combines the four projected features. Let \(a_{l,s}\) denote the learnable logit of depth \(l\) at scale \(s\). The normalized coefficient and fused feature are
\begin{equation}
\bar{\alpha}_{l,s}
=
\frac{\exp(a_{l,s})}
{\sum_{l'\in\mathcal{L}}\exp(a_{l',s})},
\qquad
F_s
=
\sum_{l\in\mathcal{L}}
\bar{\alpha}_{l,s}U_{l,s}.
\end{equation}
The coefficients are learned during localization training and normalized separately at each scale, allowing the decoder to use a different depth mixture for coarse and fine spatial evidence.

\subsection{Global-Conditioned Coarse-to-Fine Localization}

Image-level detection and pixel-level localization provide complementary views of manipulation evidence: patch features indicate where an edit may occur, whereas the CLS representation summarizes whether the image contains manipulation cues globally. When local evidence is weak, this global representation can provide a useful prior; however, using it as a hard mask gate would propagate every detection error into localization. We therefore inject CLS context through a bounded residual modulation that can adjust, but not replace, the spatial representation.

The image-level class token provides global context to the localization branch. A two-layer MLP maps \(z_{\mathrm{cls}}\) to a conditioning vector \(c\). For the two coarser scales \(s\in\{19,37\}\), bounded residual FiLM is applied as
\begin{equation}
\widetilde{F}_s
=
F_s+
\rho_s
\left(
g_s\odot F_s+r_s
\right),
\qquad
\begin{cases}
g_s=\tanh(W_{\gamma,s}c+b_{\gamma,s}),\\
r_s=\tanh(W_{\beta,s}c+b_{\beta,s}),
\end{cases}
\end{equation}
where \(\rho_{19}=1\) and \(\rho_{37}=0.25\). The FiLM projections are zero-initialized. We restrict conditioning to H19 and H37 because these coarser features provide broader context, while leaving H74 unconditioned preserves a high-resolution pathway for fine-grained spatial evidence.

Manipulated regions may span broad structures while retaining fine boundaries, making direct prediction from one resolution inadequate for both. The decoder therefore integrates the three scales in a coarse-to-fine manner. \(\widetilde{F}_{19}\) is upsampled and fused with \(\widetilde{F}_{37}\), and the resulting feature is further upsampled and fused with \(F_{74}\). Each fusion stage reduces the concatenated representation to 128 channels. Progressive upsampling then recovers resolutions \(148\), \(296\), \(592\), and \(1036\), followed by a one-channel head that predicts the manipulation logit map.

\subsection{Training and Inference}

For each training protocol, the DINOv2 encoder is first adapted for image-level manipulation classification on the corresponding training data using LoRA. The adapted encoder and inherited classification head are then frozen, and the localization-specific modules are trained on the same data with pixel-level supervision. This preserves the image-level representation while allowing the localization branch to learn how to reuse it.

Auxiliary heads supervise H19, H37, and H74 so that every spatial pathway remains directly predictive instead of relying only on the final fusion output. Let \(\ell_s\) denote the weighted sum of binary cross-entropy and soft Dice loss at output scale (s), and let \(\lambda_s\) denote the corresponding auxiliary-loss coefficient. The localization objective is
\begin{equation}
\mathcal{L}
=
\ell_{\mathrm{final}}
+\lambda_{19}\ell_{19}
+\lambda_{37}\ell_{37}
+\lambda_{74}\ell_{74}.
\end{equation}
The auxiliary and per-image class weights are fixed training hyperparameters specified in Appendix~\ref{app:details}. No classification loss is optimized during localization training; the frozen classification branch remains available for image-level detection, while its class representation provides global context to the localization decoder.

\paragraph{Optional prototype-guided refinement.}
The direct mask can still contain locally uncertain predictions even when its confident regions are reliable. We therefore study an inference-only prototype refinement applied to the direct ForenScope prediction. Using the H37 features and resized localization scores, manipulated and real prototypes are estimated from confident regions, and their cosine-similarity margin is used to adjust the final logits. The refinement is applied with the same fixed rule across datasets, requires no target-domain labels or parameter updates, and is reported as an inference-only extension of the direct prediction. Full formulation and hyperparameters are provided in Appendix~\ref{app:prototype}.

\section{Experiments}
\label{sec:experiments}

\subsection{Experimental Setup}

We evaluate \method for image-level manipulation detection and pixel-level localization under two independently trained settings. The GIFTBench setting performs both training stages on the GIFTBench training split, whereas Protocol-MVSS repeats them using only CASIAv2; the settings share neither training data nor adapted checkpoints. All OOD datasets are reserved for evaluation, and checkpoints are selected using validation data from the corresponding training domain. Unless otherwise stated, ablations follow the GIFTBench setting.

At inference, detection and localization are predicted separately; the detection decision does not gate the mask, although CLS features condition the localization branch through FiLM. Unless otherwise specified, ForenScope denotes the direct output of the selected EMA checkpoint at a fixed threshold of 0.5. Localization F1 and IoU are computed on manipulated images, while image-level ROC AUC is reported on datasets containing both real and manipulated samples. ForenScope + Proto. denotes the fixed inference-only refinement. Full implementation details are provided in Appendix~\ref{app:details}.

\subsection{Overall Performance}
Table~\ref{tab:localization-main} reports pixel-level F1 on three held-out GIFTBench composition subsets and nine OOD datasets; complete in-domain and twelve-dataset F1/IoU results are provided in Appendix~\ref{app:forenscope-results}. On the source-disjoint GIFTBench split, \method remains strong across source, semantic-target, operation, and composition views. Across the nine OOD datasets, the direct ForenScope output achieves the highest average F1 among the compared direct-output models and outperforms SparseViT on most datasets, despite dataset-specific variation.

The lower panel reports cross-dataset results under Protocol-MVSS training. ForenScope achieves the highest average F1 among the compared methods, although its relative advantage varies across individual OOD datasets. The optional prototype refinement further improves the direct prediction and produces the strongest overall performance in this setting.
\begin{table}[t]
\centering
\fontsize{6.4pt}{7.0pt}\selectfont
\setlength{\tabcolsep}{1.2pt}
\renewcommand{\arraystretch}{0.86}
\caption{Pixel-level F1 under GIFTBench and Protocol-MVSS training.}
\label{tab:localization-main}
\begin{tabular*}{\linewidth}{@{\extracolsep{\fill}}l*{13}{c}@{}}
\toprule
\multicolumn{14}{c}{\textit{GIFTBench training}} \\
\specialrule{\heavyrulewidth}{0pt}{1.2pt}
& \multicolumn{3}{c}{In-domain GIFTBench} & \multicolumn{10}{c}{Out-of-domain datasets} \\
\cmidrule(lr){2-4}\cmidrule(lr){5-14}
Method & Single & Fusion & Multi & CASIAv1 & NIST16 & IMD2020 & COVERAGE & Columbia & MISD & AutoSplice & CocoGlide & Br-Gen & Avg \\
\midrule
PSCC-Net & .610 & .538 & .658 & .320 & .189 & .271 & .339 & .649 & .641 & .856 & .567 & .563 & .4883 \\
TruFor & .837 & .816 & .819 & .642 & .164 & .490 & .344 & .801 & \textbf{.763} & .917 & .603 & .574 & .5887 \\
IML-ViT & .730 & .656 & .793 & .628 & .251 & .552 & .241 & .742 & .749 & .868 & .306 & .568 & .5450 \\
SparseViT & .862 & .796 & .850 & .633 & .200 & .543 & .430 & \textbf{.813} & \textbf{.763} & \textbf{.923} & .698 & .643 & .6273 \\
\textbf{\method} & \textbf{.891} & \textbf{.871} & .867 & .679 & .548 & .579 & .578 & .546 & .696 & .871 & .799 & .746 & .6712 \\
\method{} + Proto. & .885 & \textbf{.896} & \textbf{.867} & \textbf{.695} & \textbf{.566} & \textbf{.580} & \textbf{.592} & .564 & .726 & .873 & \textbf{.820} & \textbf{.753} & \textbf{.6855} \\
\bottomrule
\end{tabular*}

\vspace{2pt}
\begin{tabular*}{\linewidth}{@{\extracolsep{\fill}}l*{10}{c}@{}}
\toprule
\multicolumn{11}{c}{\textit{Protocol-MVSS training: external-only comparison}} \\
\specialrule{\heavyrulewidth}{0pt}{1.2pt}
Method & CASIAv1 & NIST16 & IMD2020 & COVERAGE & Columbia & MISD & AutoSplice & CocoGlide & Br-Gen & Avg \\
\midrule
PSCC-Net & .375 & .184 & .259 & .239 & .605 & .690 & \textbf{.659} & .285 & \textbf{.267} & .3959 \\
TruFor & \textbf{.721} & .321 & .323 & .424 & \textbf{.865} & \textbf{.756} & .320 & .204 & .085 & .4466 \\
IML-ViT & .718 & .291 & .322 & .438 & .747 & .710 & .221 & .211 & .079 & .4152 \\
\textbf{\method} & .681 & .524 & .563 & .518 & .454 & .632 & .276 & .467 & .191 & .4782 \\
\method{} + Proto. & .691 & \textbf{.539} & \textbf{.571} & \textbf{.543} & .487 & .658 & .290 & \textbf{.499} & .210 & \textbf{.4986} \\
\bottomrule
\end{tabular*}
\end{table}

\paragraph{Image-level detection.}
Although our architectural contributions focus on localization, ForenScope retains the image-level classifier inherited from the classification-adapted encoder. On the eight external datasets containing both real and manipulated images, the GIFTBench-adapted detector achieves an average ROC AUC of 0.935 and F1 of 0.871. The full per-dataset comparison, including the controlled CASIAv2-trained setting, is provided in Appendix~\ref{app:forenscope-results}.

\subsection{Ablation Study}
Table~\ref{tab:ablation-analysis} evaluates the principal components of ForenScope using the nine-dataset external average. The upper block contains retrained structural variants; the lower block reports the optional fixed inference-time refinement, while the equal-layer control is provided in Appendix~\ref{tab:ablation-interventions}.
\begin{table}[t]
\centering
\scriptsize
\setlength{\tabcolsep}{3pt}
\renewcommand{\arraystretch}{0.88}
\caption{Ablation of ForenScope components on the nine-dataset external average.}
\label{tab:ablation-analysis}
\begin{tabular}{lcccc cc}
\toprule
Variant & ADE Adapt. & Multi-depth & Multi-scale & CLS-FiLM & F1 & IoU \\
\midrule
\multicolumn{7}{l}{\textit{Retrained structural variants (direct output)}} \\
w/o ADE adaptation & $\times$ & \checkmark & \checkmark & \checkmark & .632 & .535 \\
F17 only & \checkmark & $\times$ & \checkmark & \checkmark & .619 & .523 \\
H37 only & \checkmark & \checkmark & $\times$ & \checkmark & .641 & .544 \\
H74 only & \checkmark & \checkmark & $\times$ & N/A & .644 & .540 \\
w/o CLS-FiLM & \checkmark & \checkmark & \checkmark & $\times$ & .659 & .567 \\
\textbf{\method} & \checkmark & \checkmark & \checkmark & \checkmark & \textbf{.671} & \textbf{.577} \\
\midrule
\multicolumn{7}{l}{\textit{Fixed-checkpoint inference refinement}} \\
\method{} + Proto. & \checkmark & \checkmark & \checkmark & \checkmark & .685 & .592 \\
\bottomrule
\end{tabular}
\end{table}

Using only F17 causes the largest structural drop, supporting the aggregation of complementary evidence across encoder depths. The H37-only and H74-only variants perform comparably to each other but both underperform the full model, supporting the combination of spatial evidence across resolutions rather than reliance on a particular single scale. The w/o-ADE result supports classification-adapted representations, while removing CLS-FiLM has a smaller effect, suggesting a modest rather than uniform benefit from explicit conditioning. Prototype refinement improves the average without retraining; the equal-layer control and additional FiLM-off, shuffled-CLS, and joint-control diagnostics are reported in Appendix~\ref{app:ablation-details}.




\section{Conclusion}

We presented \gift, a multi-axis benchmark for diagnosing image forgery localization generalization, together with \method, a detection--localization model using learned depth--scale fusion and bounded global conditioning. \gift exposes distinct transfer behavior across manipulation source, semantic target, operation, and composition while complementing controlled diagnosis with twelve-dataset external evaluation. \method retains image-level detection and improves cross-dataset localization, demonstrating how structured benchmark diagnosis can provide testable guidance for more transferable forensic models.

\section*{AI Use Statement}

Generative AI tools were used to assist with polishing the writing of the manuscript, retrieving and identifying relevant literature, and drafting portions of the paper. All AI-assisted content was reviewed, revised, and verified by the authors, including factual claims and references. The authors take full responsibility for the final content of this work.
\clearpage
\appendix

\section{Benchmark Construction and Data Specification}
\label{app:benchmark}

This appendix section specifies what GIFTBench contains, how each manipulated sample is constructed, and how split membership, masks, and provenance are recorded. It supplements the benchmark overview in the main paper with the details needed to reconstruct protocol memberships without repeating the main motivation.

\subsection{Scope, taxonomy, and task definition}

\gift is derived from ADE20K training images and their semantic annotations~\citep{zhou2019semantic}. Its purpose is to keep the real-image universe and localization target controlled while exposing several sources of distribution shift that are usually entangled in existing IFL datasets. Each manipulated image has a split-assigned primary source image and, where applicable, split-assigned donor or component sources; it is also assigned metadata for manipulation source, semantic target, editing operation, composition level, split, construction configuration, and quality-control status. Multi-Source samples additionally retain component-level masks and provenance.

The four axes are defined as follows:

\begin{table}[ht]
\centering
\scriptsize
\setlength{\tabcolsep}{2.5pt}
\renewcommand{\arraystretch}{0.88}
\caption{The four diagnostic axes recorded by \gift. A sample may participate in multiple protocol-specific groupings.}
\label{tab:benchmark-axes}
\resizebox{0.90\linewidth}{!}{%
\begin{tabular}{p{0.24\linewidth}p{0.70\linewidth}}
\toprule
Axis & Labels and meaning \\
\midrule
Manipulation source & Splicing, Copy--Move, Removal, and AI-Edit. This identifies the family of editing pipeline that produced the sample. \\
Semantic Target & Object, Component, and Background. This identifies the semantic target category or spatial scale of the manipulated content. \\
Editing operation & Add, Remove, and Replace. This identifies the intended editing objective independently of the source family. \\
Composition complexity & Single, Fusion, and Multi-Source. This identifies whether the edit is isolated, boundary-refined, or composed with other manipulation sources. \\
\bottomrule
\end{tabular}
}%
\end{table}

These labels are intentionally not interchangeable. \emph{Removal} in the source axis denotes a family of removal pipelines, whereas \emph{Remove} in the operation axis denotes an editing objective that can also be instantiated by other source families. Similarly, Fusion and Multi-Source are composition levels derived from the same source-image universe, not independent image collections. This representation allows one sample to be grouped by different axes in different diagnostic protocols while keeping its image, mask, and provenance fixed.

The released image-level label indicates whether an image is real or manipulated, while the pixel-level target is the binary mask of pixels changed by the recorded construction operation. Source-family, semantic-target, operation, and composition labels are metadata used to form controlled evaluation groups; they are not additional prediction targets in the localization experiments. The primary source identifier links an image and its mask, while donor and component provenance is retained when applicable; an image can therefore be assigned to different protocol views without changing its supervision.

\subsection{Construction pipeline}

The construction process begins with a real ADE20K image and its semantic annotations. Human experts define the taxonomy, admissible combinations, and construction constraints. A vision--language model then analyzes the scene and produces a structured manipulation plan, including the target region, source material from the assigned split, intended operation, and spatial attributes. A family-specific editing engine performs the pixel-level change, after which automatic checks, VLM-based verification, and manual review validate the image, mask, and metadata. Figure~\ref{fig:pipeline} summarizes this construction pipeline.

The semantic taxonomy and admissible group assignments are fixed before generation. Candidate selection and editing use the recorded random state for the corresponding construction job, whereas mask derivation, normalization, quality checks, and split assignment are deterministic once the construction record is fixed. Geometric transforms are applied jointly to image and mask; photometric transforms are applied only to the image. Masks are resized with nearest-neighbor interpolation and re-binarized after every geometry-changing operation. These rules prevent augmentation from changing the localization target while preserving the intended appearance variation.

\paragraph{Semantic planning and material selection.}
VLMs are used as semantic controllers rather than as the sole pixel generators. Qwen-VL and InternVL-family models~\citep{bai2025qwen25vltechnicalreport,bai2025qwen3,wang2025internvl3} identify eligible regions and contextual relations, propose or verify target regions and source materials, and assess placement, scale, occlusion, illumination, and scene compatibility where applicable. Candidate source contents are filtered for geometric completeness and contextual compatibility within the assigned split before execution. This separation lets the benchmark vary the editing engine without making the entire release an artifact of one VLM.

\paragraph{Pixel-level execution and mask generation.}
Classical compositing, traditional or learned inpainting, and diffusion-based editors perform the actual image modification. The executor is selected according to the source family and the semantic characteristics of the target. A binary mask is generated from the recorded construction operation and identifies pixels changed relative to the original source. For composed samples, masks for the individual components are retained in addition to the final union or refined mask.

\subsection{Manipulation source instantiation}

\paragraph{Splicing.}
Splicing includes object insertion or replacement and background replacement. Background edits include partial-region, whole-region, and cross-class substitutions. For object edits, the structured plan specifies the candidate object, placement, support surface, scale, lighting relation, and occlusion constraints; compatible donor material is selected from the ADE20K pool assigned to the same train/test partition as the primary source before classical compositing executes the edit. These cases vary both local blending traces and scene-level consistency requirements.

\paragraph{Copy--Move.}
Copy--Move includes background-to-background texture copying, background-to-object overwriting, whole-object relocation, superpixel-level cloning, texture extension, and the attribute subset. Candidate regions are selected according to semantic role and geometric completeness, followed by controlled transformations, color alignment, and blending. The attribute subset is categorized as Copy--Move in the release taxonomy and is included in the corresponding source-level counts.

\paragraph{Removal.}
Removal covers background-texture repair, structural-object removal, and semantic-foreground removal. The construction plan distinguishes locally repairable textures from extended structures and complete foreground instances. Traditional inpainting~\citep{telea2004image} or learned inpainting~\citep{suvorov2022resolution,li2022mat,sargsyan2023mi,jain2023keys,li2025rorem} is selected according to the target's spatial and semantic characteristics, so the source family is not restricted to one inpainting mechanism.

\paragraph{AI-Edit.}
AI-Edit samples use mask-conditioned diffusion editing driven by visually grounded, structured instructions. The plans cover subject-level, background/stuff-level, component-level, and spatial-interaction edits; each can implement Add, Remove, or Replace. Multiple diffusion and instruction-guided inpainting backends~\citep{rombach2022high,podell2023sdxl,manukyan2023hd,zhuang2024task,batifol2025flux} are used to provide variation in generative mechanisms, while semantic target and operation remain separately recorded.

\subsection{Composition construction}

Composition is not treated as a fifth manipulation source. It tests whether evidence learned from isolated edits survives two controlled changes in the same source-image setting. \textbf{Fusion} begins with a single-operation manipulation and applies latent-diffusion inpainting~\citep{rombach2022high,podell2023sdxl} only to a narrow morphological band around the existing manipulation boundary; pixels outside the band remain unchanged. The final mask includes the refined boundary band and the retained preceding manipulation, providing a matched single-operation reference.

\textbf{Multi-Source} composition combines two to four distinct single-manipulation components derived from the same real image and its assigned source pool. The visible components are placed back onto the original background with controlled spatial layouts, and their masks are merged into a final union mask. Component-level masks and provenance support analyses beyond the union-mask localization target. The composition training partitions are released for future composition-aware methods, whereas the Protocol~IV baselines in this paper train only on GIFTBench-Core and therefore measure zero-composition generalization.

\subsection{Quality assurance, masks, and provenance}

Before large-scale generation, each manipulation family and its mask-extraction procedure is checked on pilot examples for visual plausibility and mask correctness. During generation, samples whose manipulated region is below \(0.5\%\) of the image area are discarded. Candidate samples are further filtered by geometry, manipulated-area, semantic consistency, visual quality, saliency change, and near-duplicate content. AI-Edit candidates additionally undergo BRISQUE no-reference quality assessment, used for relative ranking within each batch rather than with a fixed threshold, and CLIP-based semantic similarity between the edited region and the target instruction.

All remaining samples receive VLM-based quality assessment. Classical manipulations use three-level scoring (1: superior, 2: acceptable, 3: discard), while AI-Edit samples receive type-specific binary verification of physical plausibility and image--instruction consistency. Samples with low saliency relative to the source, low visual quality, or duplicated scene content are removed. Finally, retained samples are manually inspected for incorrect edits, implausible results, and inaccurate masks. These checks prioritize reliable localization targets rather than only maximizing the number of generated images.

The released mask is a single-channel binary image with values 0 and 255. Construction-time grayscale masks are normalized at release time using threshold 128: values \(\geq 128\) are encoded as 255 and all other values as 0. Positive pixels denote changes relative to the original source under the recorded construction procedure. The construction pipeline initially produced 115,025 candidates; twelve Copy--Move training candidates with no positive pixels after normalization were excluded as invalid localization targets, yielding the final release of 115,013 samples. The excluded records are documented in the release audit and are not treated as a separate benchmark category.

Each release record stores the manipulated-image identifier, original source identifier, mask, four-axis labels, split, construction engine and configuration identifier, and quality-control decision. For Multi-Source samples, component masks and component-level provenance are also retained. This record structure allows protocol memberships and dataset statistics to be regenerated from a common manifest.

\subsection{Dataset composition and split integrity}

Before manipulation generation, all ADE20K primary and donor source images are assigned to one of two disjoint pools, train or test. Every source image and all donor/component material used to construct a sample come from the same pool; the generated manipulation and all of its descendants inherit that assignment. Thus, no primary source, donor source, source material, or manipulated descendant is shared across the train/test boundary. The complete release composition is:

\begin{table}[ht]
\caption{Release composition of \gift. GIFTBench-Core is the single-operation benchmark used for baseline training; composition training partitions are released for future composition-aware methods. The attribute subset is included under Copy--Move.}
\label{tab:release_counts}
\centering
\scriptsize
\setlength{\tabcolsep}{2.5pt}
\renewcommand{\arraystretch}{0.90}
\begin{tabular}{lrr}
\toprule
Release partition & Train & Test \\
\midrule
\multicolumn{3}{l}{\textit{GIFTBench-Core (single-operation)}} \\
Splicing & 20,673 & 2,327 \\
Copy--Move (including attribute) & 20,887 & 2,386 \\
Removal & 21,074 & 2,380 \\
AI-Edit & 25,758 & 2,851 \\
\midrule
Core subtotal & 88,392 & 9,944 \\
\midrule
\multicolumn{3}{l}{\textit{GIFTBench-Composition (challenge set)}} \\
Fusion & 5,925 & 750 \\
Multi-Source & 9,002 & 1,000 \\
\midrule
Total & 103,319 & 11,694 \\
\bottomrule
\end{tabular}
\end{table}

The final release contains 115,013 manipulated images: 103,319 training and 11,694 test samples. The split definitions and protocol memberships in the machine-readable manifest are the source of truth for these counts. The release distributes the manipulated images and masks but not the original ADE20K source images; users obtain the latter independently under their original terms. This separation preserves the applicable source-data conditions while allowing users with access to ADE20K to reconstruct paired provenance for auditing.

\subsection{Dataset-level visual summaries}

\begin{figure}[H]
    \centering
    \begin{minipage}[t]{0.49\textwidth}
        \centering
        \includegraphics[width=\linewidth]{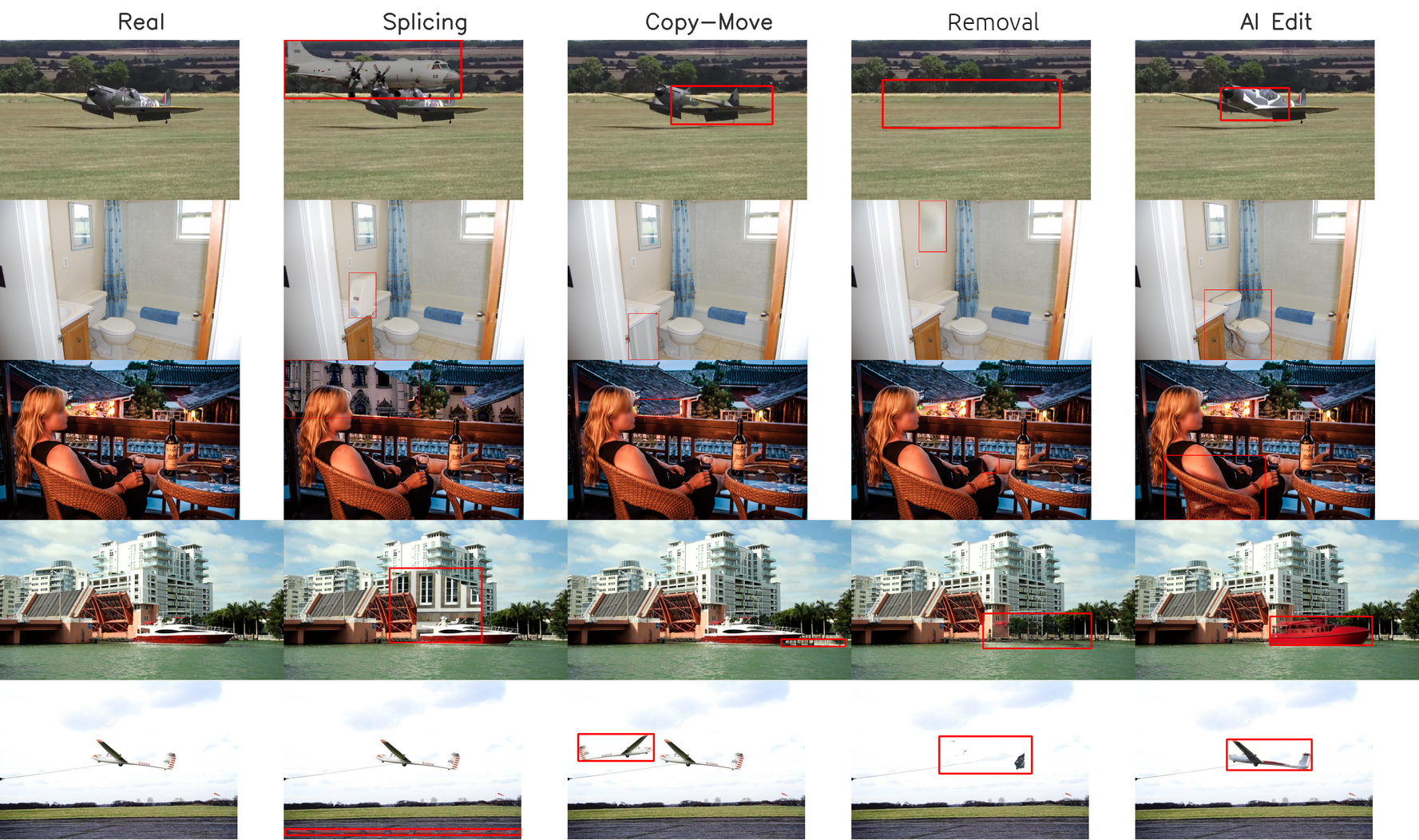}
        \par
        {\footnotesize (a) Representative construction examples\par}
    \end{minipage}\hfill
    \begin{minipage}[t]{0.49\textwidth}
        \centering
        \includegraphics[width=\linewidth]{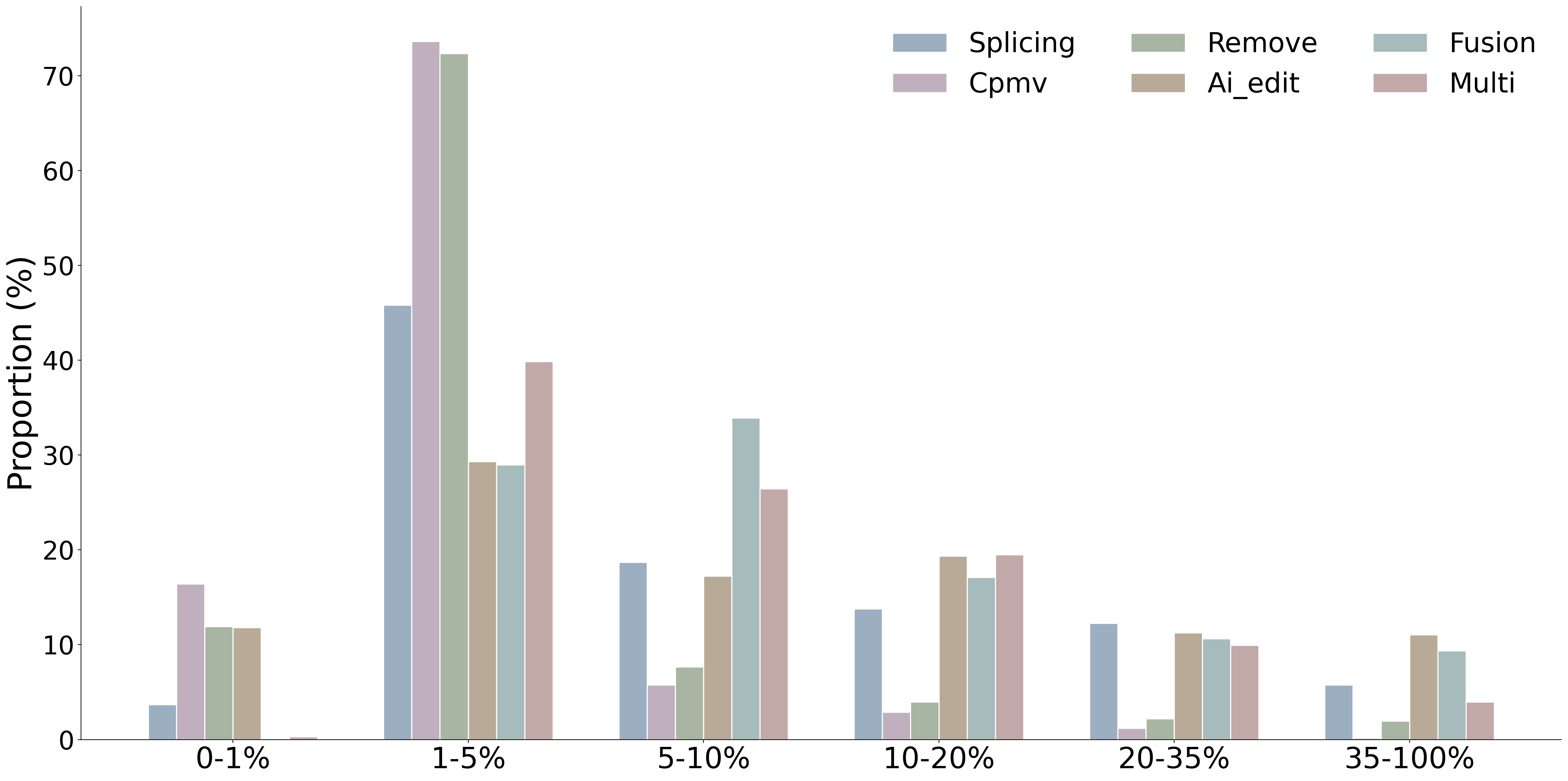}
        \par
        {\footnotesize (b) Manipulated-area distributions\par}
    \end{minipage}
    \caption{Dataset-level summaries of GIFTBench: representative manipulation-source examples and manipulated-area distributions across source and composition groups.}
    \label{fig:dataset-visual-summaries}
\end{figure}

The release statistics are complemented by two visual summaries. Figure~\ref{fig:dataset-visual-summaries}(a) gives representative instances of the four manipulation-source families, while Fig.~\ref{fig:dataset-visual-summaries}(b) shows manipulated-area distributions across the source and composition groups. These views describe the released benchmark rather than additional evaluation results.

\section{Benchmark Protocols and Diagnosis-to-Method Bridge}
\label{app:diagnosis}

This section fixes the interpretation of every benchmark result and makes explicit how the diagnosis motivates the ForenScope design. It distinguishes public-checkpoint transfer, controlled retraining, and GIFTBench-to-external transfer before giving the metric conventions and the supplemental protocol profiles.

\subsection{Protocol definitions}

Table~\ref{tab:appendix-protocols} summarizes the four controlled protocols and the two transfer settings that appear throughout the paper. A training source in this table denotes the data or checkpoint used to obtain the model; it does not imply that the test partition is used for adaptation.

\begin{table}[t]
\centering
\scriptsize
\setlength{\tabcolsep}{2.5pt}
\renewcommand{\arraystretch}{0.90}
\begin{tabular}{p{0.17\textwidth}p{0.23\textwidth}p{0.22\textwidth}p{0.28\textwidth}}
\toprule
\textbf{Setting} & \textbf{Training source} & \textbf{Evaluation partition} & \textbf{Diagnostic question} \\
\midrule
\multicolumn{4}{l}{\textit{Axis-specific GIFTBench protocols}} \\
Protocol I & One source group & All four source groups & Source transfer: does evidence transfer across manipulation families? \\
Protocol II & One semantic target group & All three semantic target groups & Semantic-target transfer: does behavior survive changes in target category? \\
Protocol III & One operation group & All three operation groups & Operation transfer: does evidence transfer across editing intent? \\
Protocol IV & Single-operation training & Single, Fusion, and Multi-Source groups & Composition: does isolated-edit evidence survive composition? \\
\midrule
\multicolumn{4}{l}{\textit{Cross-domain transfer settings}} \\
Public checkpoint transfer & Released checkpoint, no adaptation & GIFTBench evaluation groups & What is the training-distribution gap of existing models? \\
External transfer & GIFTBench-trained model & Twelve external datasets & Does benchmark training transfer beyond its construction domain? \\
\bottomrule
\end{tabular}
\caption{Benchmark evaluation settings. The first block organizes training and testing according to one focal GIFTBench axis at a time; the second block distinguishes public-checkpoint transfer to GIFTBench from GIFTBench-trained transfer to external datasets.}
\label{tab:appendix-protocols}
\end{table}

\subsection{Metric and aggregation definitions}
\label{app:metric-aggregation}

For each manipulated image, the predicted probability map is thresholded at 0.5 and compared with the binary target mask. Pixel-level F1 and IoU are computed independently for each manipulated image and then averaged within each dataset or benchmark group. Real images are not assigned a manipulated-mask F1/IoU when no positive mask exists; their predicted-positive area is retained only for image-level false-positive inspection. Cross-dataset summaries macro-average the dataset-level scores, so All-12 is the equal-weight mean over the twelve external datasets and the Avg columns in the main tables are equal-weight means over the displayed datasets.

Image-level detection uses the frozen CLS head score and ROC AUC. AUC is defined only when both real and manipulated images are present in the evaluated split; the four pure-manipulated external splits therefore receive ``--'' rather than a fabricated AUC. The image-level score and localization mask are retained as separate outputs at inference; no post-hoc detector gate is applied to the mask. Unless a table states otherwise, localization uses the EMA checkpoint, the fixed threshold above, and the direct ForenScope output. Optional ForenScope + Proto. results apply the fixed prototype refinement described in Appendix~\ref{app:prototype}.

\subsection{Diagnosis-to-design mapping}

The benchmark analysis is descriptive: it exposes recurring transfer patterns, from which we formulate architectural hypotheses. Table~\ref{tab:diagnosis-design-map} records the intended mapping without treating the observations as causal proof.

\begin{table}[t]
\centering
\tiny
\setlength{\tabcolsep}{2.5pt}
\renewcommand{\arraystretch}{0.88}
\resizebox{0.92\linewidth}{!}{%
\begin{tabular}{p{0.22\textwidth}p{0.25\textwidth}p{0.22\textwidth}p{0.22\textwidth}}
\toprule
\textbf{Benchmark observation} & \textbf{Evidence in benchmark analysis} & \textbf{Design requirement / hypothesis} & \textbf{Architectural realization} \\
\midrule
Source-transfer asymmetry & Protocol~I matrices show strong dependence on the manipulation source & Avoid a single source-sensitive representation; test whether complementary evidence reduces this specialization & Multi-depth patch features \\
Transfer profiles vary across inputs and groups & Aggregate profiles are not uniform across examples or conditions & Combine complementary depth evidence with scale-specific learnable aggregation & Learnable scale-wise LayerMixer \\
Composition creates heterogeneous spatial evidence & Fusion/Multi-Source results and attention analysis show changes in extent, boundaries, and evidence distribution & Represent coarse extent and fine boundaries at multiple resolutions & H19/H37/H74 coarse-to-fine pathway \\
Recall-dominated errors under source shift & False-negative decomposition shows that unfamiliar local evidence is often missed & Test whether image-level manipulation context can complement ambiguous local evidence & Bounded CLS-conditioned FiLM on H19/H37 \\
\textit{Resolution-specific architectural choice} & The benchmark does not identify a unique conditioning location & Preserve a high-resolution local pathway while conditioning broader-context features & No FiLM on H74 \\
\bottomrule
\end{tabular}
}%
\caption{Diagnosis-to-design mapping used to formulate ForenScope. The table separates benchmark observations from intermediate design requirements and their architectural realization; the mapping expresses testable motivation rather than a causal conclusion.}
\label{tab:diagnosis-design-map}
\end{table}

The table intentionally separates benchmark-derived requirements from implementation choices. In particular, the absence of FiLM at H74 is a resolution-specific architectural decision evaluated by ablation, not a direct conclusion from the benchmark. The ablations in Appendix~\ref{app:forenscope-results} test whether the corresponding realizations improve transfer, while the benchmark analysis alone is not presented as causal proof.

\subsection{Supplemental diagnostic visualizations}

\begin{figure}[t]
    \centering
    \begin{minipage}[t]{0.34\textwidth}
        \centering
        \includegraphics[height=1.45in,keepaspectratio]{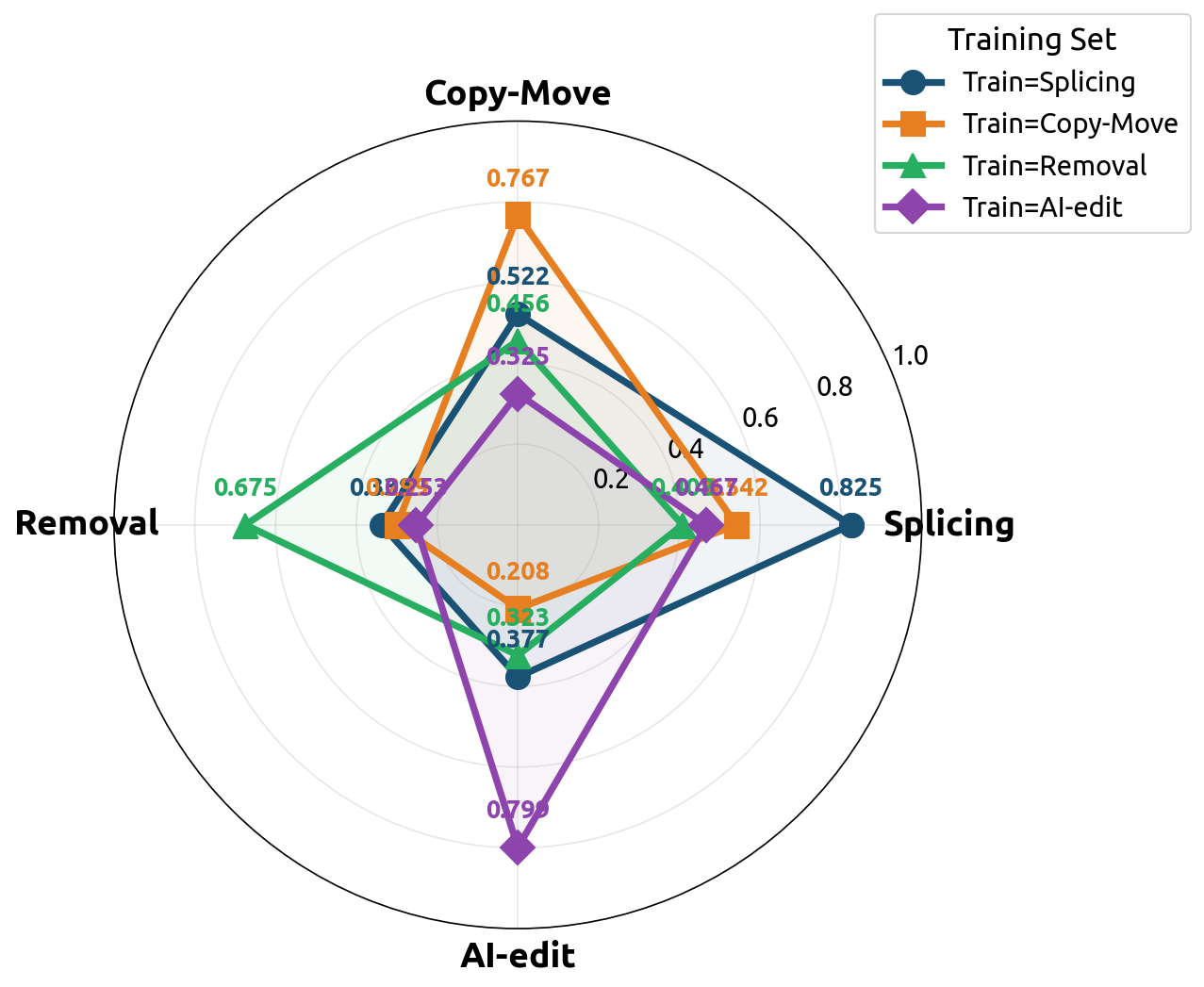}
        \par
        {\footnotesize (a) Source transfer\par}
    \end{minipage}\hfill
    \begin{minipage}[t]{0.62\textwidth}
        \centering
        \includegraphics[height=1.45in,keepaspectratio]{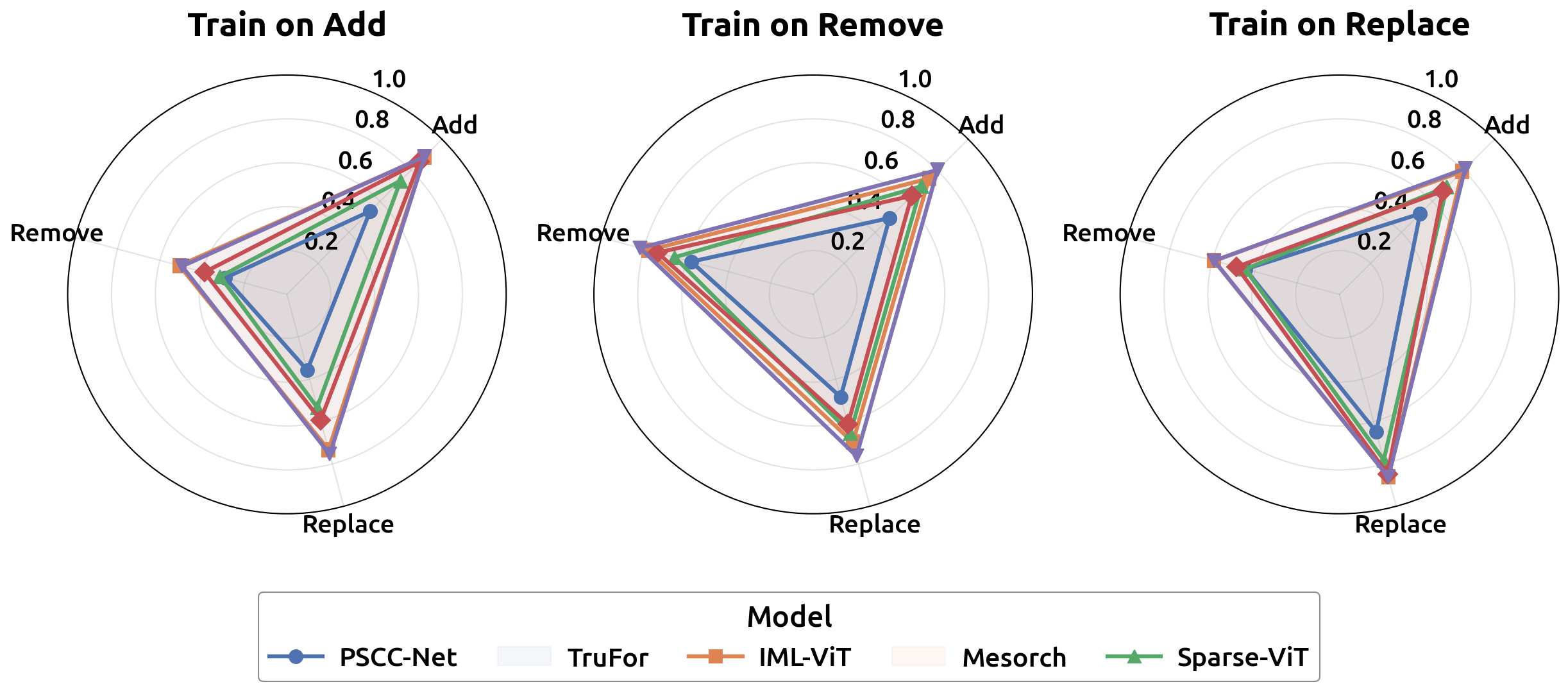}
        \par
        {\footnotesize (b) Operation transfer\par}
    \end{minipage}
    \begin{minipage}[t]{0.96\textwidth}
        \centering
        \includegraphics[width=\linewidth,height=1.75in,keepaspectratio]{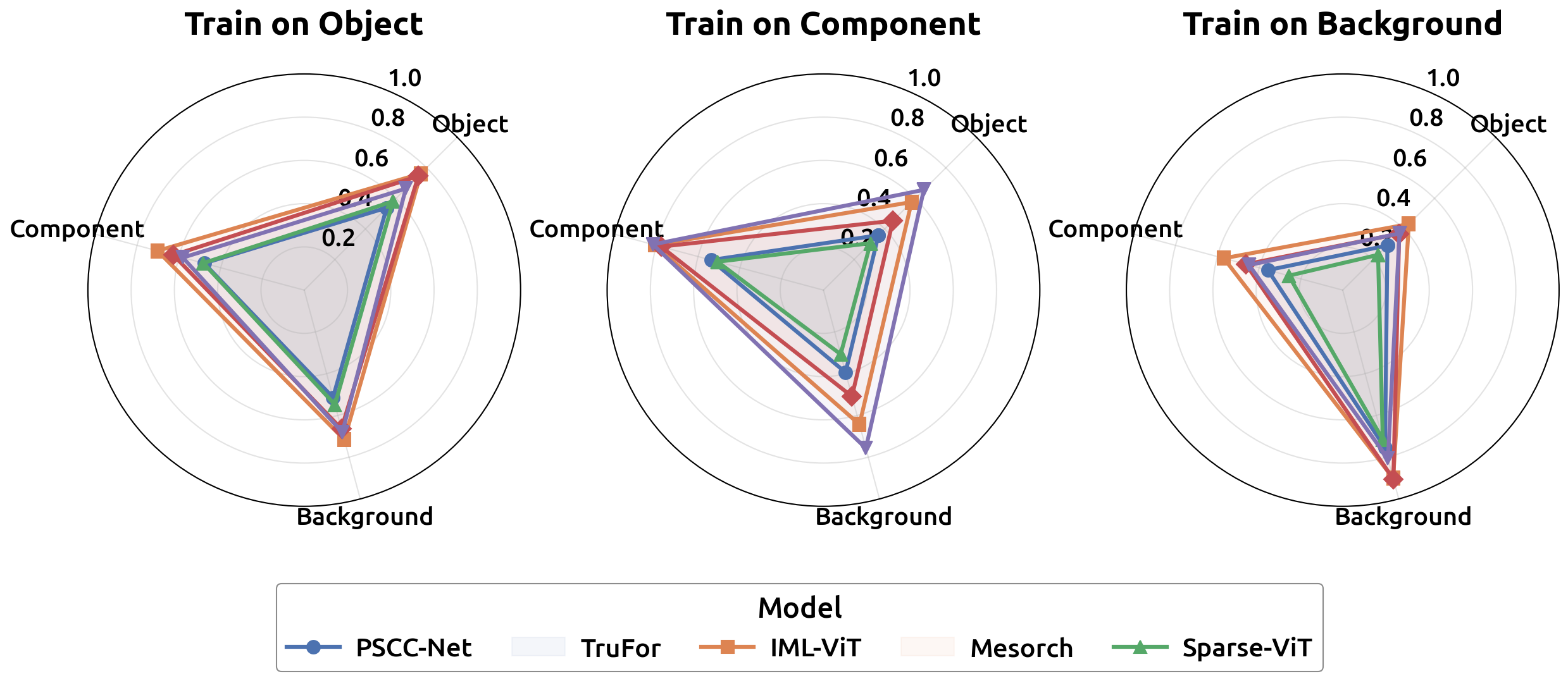}
        \par
        {\footnotesize (c) Cross-semantic-target transfer profile\par}
    \end{minipage}
    \caption{Aggregate transfer profiles for the benchmark protocols. Panels (a)--(c) summarize cross-source, cross-operation, and cross-semantic-target localization performance, respectively. These radar plots complement the fine-grained diagnostic visualizations in the main text and are intended as descriptive summaries rather than causal evidence.}
    \label{fig:diagnosis-details}
\end{figure}

The main text presents detailed source, semantic-target, operation, and composition diagnostics. Figure~\ref{fig:diagnosis-details} provides complementary aggregate radar profiles for Protocols~I--III. Panels (a)--(c) summarize cross-source, cross-operation, and cross-semantic-target localization behavior; they are descriptive comparisons and do not establish causal explanations for the observed gaps.

\section{Complete Benchmark Results and Robustness}
\label{app:benchmark-results}

This section provides the complete controlled benchmark matrices and the available stress-test evidence. Each table states its training and test convention explicitly so that diagonal performance, cross-condition transfer, and composition references are not conflated.

\subsection{Released-checkpoint transfer to GIFTBench}

Table~\ref{tab:giftbench-public-transfer} reports the complete evaluation of existing released or unified checkpoints on GIFTBench without additional training or fine-tuning. The training-protocol column identifies the source protocol of each checkpoint. Because these models were trained on different data, the table measures transfer to GIFTBench rather than a matched architectural comparison. ForenScope is reported separately in Table~\ref{tab:localization-main}, since its main configuration is trained on GIFTBench and is therefore not comparable to these zero-adaptation rows.

\begin{table}[H]
\centering
\tiny
\setlength{\tabcolsep}{1.3pt}
\renewcommand{\arraystretch}{0.86}
\caption{Released-checkpoint transfer to GIFTBench without adaptation. Entries are pixel-level F1/IoU.}
\label{tab:giftbench-public-transfer}
\begin{tabular*}{0.94\textwidth}{@{\extracolsep{\fill}}ll*{14}{c}@{}}
\toprule
\multirow{3}{*}{\shortstack{Training\\Protocol}} & \multirow{3}{*}{Model}
& \multicolumn{8}{c}{Manipulation Source}
& \multicolumn{6}{c}{Composition} \\
\cmidrule(lr){3-10}\cmidrule(lr){11-16}
&& \multicolumn{2}{c}{Splicing} & \multicolumn{2}{c}{Copy--Move}
& \multicolumn{2}{c}{Removal} & \multicolumn{2}{c}{AI--Edit}
& \multicolumn{2}{c}{Single} & \multicolumn{2}{c}{Fusion}
& \multicolumn{2}{c}{Multi--Source} \\
\cmidrule(lr){3-4}\cmidrule(lr){5-6}\cmidrule(lr){7-8}\cmidrule(lr){9-10}
\cmidrule(lr){11-12}\cmidrule(lr){13-14}\cmidrule(lr){15-16}
&& F1 & IoU & F1 & IoU & F1 & IoU & F1 & IoU & F1 & IoU & F1 & IoU & F1 & IoU \\
\midrule
\multirow{6}{*}{Protocol-MVSS}
& CAT-Net & .376 & .293 & .239 & .171 & .128 & .084 & .151 & .106 & .128 & .084 & .151 & .106 & .219 & .160 \\
& ObjectFormer & .170 & .106 & .077 & .043 & .070 & .041 & .168 & .105 & .070 & .041 & .168 & .105 & .174 & .103 \\
& PSCC-Net & .295 & .209 & .161 & .103 & .074 & .045 & .143 & .094 & .074 & .045 & .143 & .094 & .233 & .149 \\
& MVSS-Net & .307 & .224 & .219 & .152 & .105 & .067 & .142 & .095 & .105 & .067 & .142 & .095 & .286 & .194 \\
& TruFor & .591 & .533 & .490 & .426 & .235 & .195 & .199 & .162 & .235 & .195 & .199 & .162 & .450 & .353 \\
& IML-ViT & .548 & .482 & .448 & .373 & .191 & .150 & .180 & .139 & .191 & .150 & .180 & .139 & .452 & .358 \\
\midrule
\multirow{11}{*}{Protocol-CAT}
& MVSS-Net & .402 & .318 & .246 & .177 & .134 & .090 & .152 & .104 & .134 & .090 & .152 & .104 & .230 & .170 \\
& CAT-Net & .556 & .494 & .392 & .330 & .210 & .167 & .133 & .107 & .210 & .167 & .133 & .107 & .430 & .340 \\
& ObjectFormer & .207 & .155 & .124 & .087 & .051 & .033 & .104 & .073 & .121 & .081 & .112 & .073 & .175 & .118 \\
& PSCC-Net & .394 & .303 & .254 & .182 & .133 & .091 & .156 & .106 & .133 & .091 & .156 & .106 & .332 & .234 \\
& TruFor & .607 & .557 & .479 & .422 & .265 & .228 & .219 & .186 & .265 & .228 & .219 & .186 & .479 & .390 \\
& IML-ViT & .622 & .566 & .514 & .441 & .236 & .197 & .194 & .155 & .236 & .197 & .194 & .155 & .534 & .446 \\
& Mesorch & .620 & .566 & .480 & .419 & .267 & .227 & .229 & .191 & .267 & .227 & .229 & .191 & .483 & .389 \\
& SparseViT & .640 & .583 & .488 & .418 & .249 & .205 & .232 & .188 & .249 & .205 & .232 & .188 & .506 & .410 \\
& SAFIRE & .329 & .266 & .177 & .134 & .125 & .091 & .190 & .143 & .203 & .157 & .248 & .198 & .293 & .217 \\
& DetectiveSAM & .164 & .118 & .109 & .069 & .084 & .051 & .093 & .063 & .084 & .051 & .093 & .063 & .158 & .100 \\
& ForensicsSAM & \textbf{.671} & \textbf{.619} & \textbf{.598} & \textbf{.537} & \textbf{.281} & \textbf{.240} & \textbf{.304} & \textbf{.259} & .281 & .240 & .304 & .259 & \textbf{.544} & \textbf{.444} \\
\midrule
MMTD-Set & FakeShield & .597 & .519 & .215 & .170 & .161 & .118 & .302 & .246 & \textbf{.330} & \textbf{.274} & \textbf{.414} & \textbf{.341} & .371 & .282 \\
\bottomrule
\end{tabular*}
\end{table}

\subsection{Protocol I complete matrix}

The rows in Table~\ref{tab:intra-domain-p1} fix the training source/condition and the columns enumerate all four test sources. Each cell reports F1/IoU/AUC; the AUC is retained because these protocol splits include the corresponding image-level classification labels. The Copy--Move block includes the attribute-containing subset under the same source label.

\begin{table}[H]
  \centering
  \tiny
  \renewcommand{\arraystretch}{0.88}
  \setlength{\tabcolsep}{1.8pt}

  \begin{tabular*}{0.94\textwidth}{@{\extracolsep{\fill}}llcccccccccccc}

    \toprule

    \multirow{2}{*}{\textbf{Training Set}}
    &
    \multirow{2}{*}{\textbf{Model}}
    &
    \multicolumn{3}{c}{\textbf{Splicing}}
    &
    \multicolumn{3}{c}{\textbf{Copy-Move}}
    &
    \multicolumn{3}{c}{\textbf{Removal}}
    &
    \multicolumn{3}{c}{\textbf{AI-Edit}}
    \\

    \cmidrule(lr){3-5}
    \cmidrule(lr){6-8}
    \cmidrule(lr){9-11}
    \cmidrule(lr){12-14}

    &
    &
    F1 & IoU & AUC
    &
    F1 & IoU & AUC
    &
    F1 & IoU & AUC
    &
    F1 & IoU & AUC
    \\

    \midrule

    \multirow{5}{*}{\textbf{Splicing}}

      & PSCC-Net 
      & 0.5143 & 0.4149 & 0.9022
      & 0.2333 & 0.1590 & 0.8254
      & 0.1723 & 0.1184 & 0.7189
      & 0.2258 & 0.1686 & 0.6948
      \\

      & TruFor 
      & 0.9024 & 0.8524 & \underline{0.9951}
      & 0.5732 & 0.4961 & \underline{0.9440}
      & \underline{0.4178} & \underline{0.3565} & \underline{0.8537}
      & \underline{0.4870} & \underline{0.4183} & \underline{0.8925}
      \\

      & IML-ViT 
      & 0.8753 & 0.8443 & 0.9262
      & \underline{0.5931} & \underline{0.5275} & 0.8454
      & 0.3456 & 0.2959 & 0.7506
      & 0.3500 & 0.2962 & 0.7655
      \\

      & Mesorch 
      & \underline{0.9155} & 0.8733 & 0.9935
      & 0.5380 & 0.4795 & 0.9293
      & 0.3130 & 0.2683 & 0.8054
      & 0.2269 & 0.1896 & 0.7679
      \\

      & SparseViT 
      & \textbf{0.9187} & \underline{0.8703} & \textbf{0.9972}
      & \textbf{0.6730} & \textbf{0.5996} & \textbf{0.9717}
      & \textbf{0.4392} & \textbf{0.3791} & \textbf{0.8663}
      & \textbf{0.5964} & \textbf{0.5308} & \textbf{0.9247}
      \\

    \midrule

    \multirow{5}{*}{\textbf{Copy-Move}}

      & PSCC-Net 
      & 0.4657 & 0.3602 & 0.8710
      & 0.5583 & 0.4465 & 0.9625
      & 0.2604 & 0.1897 & 0.7732
      & 0.2153 & 0.1513 & 0.7246
      \\

      & TruFor 
      & \underline{0.5869} & \underline{0.5089} & \underline{0.9394}
      & \underline{0.8625} & \textbf{0.7981} & \underline{0.9942}
      & \underline{0.3385} & \underline{0.2958} & \underline{0.7922}
      & \underline{0.2261} & \underline{0.1791} & \underline{0.8021}
      \\

      & IML-ViT 
      & 0.4324 & 0.3592 & 0.8239
      & 0.7067 & 0.6510 & 0.8655
      & 0.2306 & 0.1915 & 0.6724
      & 0.1401 & 0.1052 & 0.6754
      \\

      & Mesorch 
      & 0.5011 & 0.4331 & 0.8994
      & 0.8406 & 0.7788 & 0.9865
      & 0.2919 & 0.2530 & 0.7581
      & 0.1201 & 0.0939 & 0.7105
      \\

      & SparseViT 
      & \textbf{0.7224} & \textbf{0.6495} & \textbf{0.9680}
      & \textbf{0.8641} & \underline{0.7965} & \textbf{0.9958}
      & \textbf{0.3738} & \textbf{0.3247} & \textbf{0.8209}
      & \textbf{0.3373} & \textbf{0.2799} & \textbf{0.8509}
      \\
    \midrule


    \multirow{5}{*}{\textbf{Removal}}

      & PSCC-Net 
      & 0.3500 & 0.2596 & 0.8066
      & 0.3043 & 0.2190 & 0.8609
      & 0.5780 & 0.4821 & 0.9406
      & 0.3153 & 0.2356 & 0.8048
      \\

      & TruFor 
      & 0.4533 & 0.3831 & \underline{0.8849}
      & 0.5258 & 0.4506 & \underline{0.9374}
      & \underline{0.7586} & \underline{0.6896} & \underline{0.9813}
      & \underline{0.4063} & \underline{0.3314} & \underline{0.8943}
      \\

      & IML-ViT 
      & \underline{0.4971} & \underline{0.4249} & 0.8313
      & \underline{0.5570} & \underline{0.4807} & 0.8415
      & 0.6120 & 0.5562 & 0.8567
      & 0.2543 & 0.2035 & 0.7849
      \\

      & Mesorch 
      & 0.2009 & 0.1579 & 0.7758
      & 0.3303 & 0.2829 & 0.8528
      & 0.6412 & 0.5872 & 0.9280
      & 0.1054 & 0.0763 & 0.7164
      \\

      & SparseViT 
      & \textbf{0.5367} & \textbf{0.4624} & \textbf{0.9226}
      & \textbf{0.5644} & \textbf{0.4814} & \textbf{0.9547}
      & \textbf{0.7829} & \textbf{0.7118} & \textbf{0.9863}
      & \textbf{0.5359} & \textbf{0.4514} & \textbf{0.9316}
      \\

    \midrule

    \multirow{5}{*}{\textbf{AI-Edit}}

      & PSCC-Net 
      & \textbf{0.6130} & \textbf{0.5190} & \textbf{0.8901}
      & \textbf{0.4392} & \textbf{0.3473} & 0.8609
      & \textbf{0.3485} & \underline{0.2700} & 0.8039
      & 0.7930 & 0.7071 & 0.9790
      \\

      & TruFor 
      & \underline{0.5203} & \underline{0.4579} & \underline{0.8807}
      & \underline{0.3862} & \underline{0.3223} & \textbf{0.8950}
      & \underline{0.3380} & \textbf{0.2825} & \textbf{0.8354}
      & \underline{0.8507} & \underline{0.7827} & \underline{0.9914}
      \\

      & IML-ViT 
      & 0.3041 & 0.2583 & 0.7309
      & 0.2025 & 0.1629 & 0.7250
      & 0.1305 & 0.1011 & 0.6649
      & 0.7344 & 0.6683 & 0.9147
      \\

      & Mesorch 
      & 0.4075 & 0.3530 & 0.8402
      & 0.3493 & 0.2956 & \underline{0.8802}
      & 0.2485 & 0.1995 & 0.7988
      & \textbf{0.8737} & \textbf{0.8116} & \textbf{0.9928}
      \\

      & SparseViT 
      & 0.4912 & 0.4235 & 0.8532
      & 0.2483 & 0.1984 & 0.8272
      & 0.1978 & 0.1535 & 0.7634
      & 0.7434 & 0.6666 & 0.9710
      \\

    \bottomrule

  \end{tabular*}
  \caption{Complete Protocol~I cross-source evaluation results (F1/IoU/AUC). Each block groups models by training source and evaluates all four test sources; all Copy--Move variants are included in the Copy--Move view.}
  \label{tab:intra-domain-p1}
\end{table}

The diagonal entries in this matrix measure within-source performance, whereas the off-diagonal entries measure transfer to a different manipulation source. The diagonal-to-off-diagonal variation is the reason the paper reports the matrix rather than only a single benchmark average.

\subsection{Protocol II complete matrix}

Table~\ref{tab:intra-domain-p2} groups models by the semantic target used for training and evaluates object-, component-, and background-level test target groups. The source and operation distributions remain part of the benchmark construction; this protocol organizes the Semantic Target view without claiming a one-variable-at-a-time intervention.

\begin{table}[t]
  \centering
  \tiny
  \renewcommand{\arraystretch}{0.88}
  \setlength{\tabcolsep}{1.8pt}
  \begin{tabular*}{0.94\textwidth}{@{\extracolsep{\fill}}llccccccccc}

    \toprule

    \textbf{Training Set}
    &
    \textbf{Model}
    &
    \multicolumn{3}{c}{\textbf{Object}}
    &
    \multicolumn{3}{c}{\textbf{Component}}
    &
    \multicolumn{3}{c}{\textbf{Background}}
    \\

    \cmidrule(lr){3-5}
    \cmidrule(lr){6-8}
    \cmidrule(lr){9-11}

    &
    &
    F1 & IoU & AUC
    &
    F1 & IoU & AUC
    &
    F1 & IoU & AUC
    \\

    \midrule


    \multirow{5}{*}{\textbf{Object}}

      & PSCC-Net
      & 0.4452 & 0.3530 & 0.9078
      & 0.3887 & 0.2895 & 0.8970
      & 0.4244 & 0.3330 & 0.8357
      \\

      & TruFor
      & \underline{0.7161} & \underline{0.6466} & \underline{0.9723}
      & \underline{0.6683} & \underline{0.5794} & \underline{0.9638}
      & \underline{0.7125} & \underline{0.6440} & \underline{0.9484}
      \\

      & IML-ViT
      & 0.6497 & 0.5948 & 0.8676
      & 0.5595 & 0.4882 & 0.8452
      & 0.6110 & 0.5481 & 0.8491
      \\

      & Mesorch
      & 0.6478 & 0.5807 & 0.9575
      & 0.5266 & 0.4465 & 0.9280
      & 0.5345 & 0.4677 & 0.9009
      \\

      & SparseViT
      & \textbf{0.7793} & \textbf{0.7077} & \textbf{0.9861}
      & \textbf{0.7019} & \textbf{0.6119} & \textbf{0.9733}
      & \textbf{0.7551} & \textbf{0.6885} & \textbf{0.9665}
      \\

    \midrule

    \multirow{5}{*}{\textbf{Component}}

      & PSCC-Net
      & 0.3595 & 0.2697 & 0.8668
      & 0.5382 & 0.4286 & 0.9407
      & 0.3942 & 0.2992 & 0.8206
      \\

      & TruFor
      & \underline{0.5760} & \underline{0.5035} & \underline{0.9364}
      & \underline{0.8080} & \underline{0.7275} & \underline{0.9877}
      & \underline{0.6435} & \underline{0.5765} & \underline{0.9220}
      \\

      & IML-ViT
      & 0.3069 & 0.2591 & 0.7770
      & 0.5067 & 0.4410 & 0.8481
      & 0.3066 & 0.2574 & 0.7180
      \\

      & Mesorch
      & 0.4542 & 0.3909 & 0.8922
      & 0.7767 & 0.7011 & 0.9774
      & 0.5093 & 0.4455 & 0.8809
      \\

      & SparseViT
      & \textbf{0.6563} & \textbf{0.5837} & \textbf{0.9572}
      & \textbf{0.8159} & \textbf{0.7371} & \textbf{0.9895}
      & \textbf{0.7569} & \textbf{0.6911} & \textbf{0.9685}
      \\
    \midrule

    \multirow{5}{*}{\textbf{Background}}

      & PSCC-Net
      & 0.2921 & 0.2354 & 0.7738
      & 0.3583 & 0.2824 & 0.8328
      & 0.7550 & 0.6783 & 0.9658
      \\

      & TruFor
      & \textbf{0.4318} & \textbf{0.3762} & \textbf{0.8595}
      & \textbf{0.5693} & \textbf{0.4929} & \textbf{0.9179}
      & \underline{0.9009} & \underline{0.8534} & \textbf{0.9947}
      \\

      & IML-ViT
      & 0.2312 & 0.1931 & 0.6992
      & 0.2580 & 0.2115 & 0.7247
      & 0.7154 & 0.6617 & 0.8905
      \\

      & Mesorch
      & 0.3690 & \underline{0.3213} & 0.8371
      & \underline{0.4634} & \underline{0.3973} & 0.8892
      & \textbf{0.9052} & \textbf{0.8634} & \underline{0.9920}
      \\

      & SparseViT
      & \underline{0.3713} & 0.3161 & \underline{0.8494}
      & 0.4481 & 0.3692 & \underline{0.9100}
      & 0.8070 & 0.7407 & 0.9859
      \\

    \bottomrule

  \end{tabular*}
 \caption{Complete Protocol~II cross-semantic-target evaluation results (F1/IoU/AUC). Each block groups models by training semantic target and evaluates all three test target groups.}
  \label{tab:intra-domain-p2}
\end{table}

The matrix reports transfer patterns associated with semantic-target group changes under a shared construction and evaluation procedure. It does not claim to statistically remove source-related or operation-related variation, and should not be read as fully separating semantic-target shift from source shift.

\subsection{Protocol III complete matrix}

Table~\ref{tab:intra-domain-p3} groups models by the editing operation used for training and evaluates Add, Remove, and Replace. It complements Protocols~I and II by examining operation-associated transfer under the same source-image universe and pixel-level task definition, without assuming that the remaining empirical distributions are statistically matched.

\begin{table}[t]
  \centering
  \scriptsize
  \renewcommand{\arraystretch}{0.95}
  \setlength{\tabcolsep}{3pt}
  \begin{tabular*}{\textwidth}{@{\extracolsep{\fill}}llccccccccc}

    \toprule

    \textbf{Training Set}
    &
    \textbf{Model}
    &
    \multicolumn{3}{c}{\textbf{Add}}
    &
    \multicolumn{3}{c}{\textbf{Remove}}
    &
    \multicolumn{3}{c}{\textbf{Replace}}
    \\

    \cmidrule(lr){3-5}
    \cmidrule(lr){6-8}
    \cmidrule(lr){9-11}

    &
    &
    F1 & IoU & AUC
    &
    F1 & IoU & AUC
    &
    F1 & IoU & AUC
    \\

    \midrule


    \multirow{5}{*}{\textbf{Add}}

      & PSCC-Net
      & 0.5362 & 0.4318 & 0.9407
      & 0.2908 & 0.2139 & 0.8105
      & 0.3581 & 0.2709 & 0.8263
      \\

      & TruFor
      & \underline{0.8861} & \textbf{0.8316} & \underline{0.9951}
      & \textbf{0.5083} & \textbf{0.4422} & \textbf{0.8842}
      & \underline{0.7327} & \underline{0.6630} & \underline{0.9616}
      \\

      & IML-ViT
      & 0.7332 & 0.6820 & 0.8741
      & 0.3151 & 0.2651 & 0.7519
      & 0.5335 & 0.4695 & 0.8559
      \\

      & Mesorch
      & 0.8551 & 0.7998 & 0.9883
      & 0.3901 & 0.3309 & 0.8408
      & 0.5927 & 0.5215 & 0.9282
      \\

      & SparseViT
      & \textbf{0.8878} & \underline{0.8309} & \textbf{0.9956}
      & \underline{0.4947} & \underline{0.4301} & \underline{0.8826}
      & \textbf{0.7550} & \textbf{0.6874} & \textbf{0.9683}
      \\

    \midrule

    \multirow{5}{*}{\textbf{Remove}}

      & PSCC-Net
      & 0.4398 & 0.3465 & 0.8991
      & 0.5328 & 0.4380 & 0.9187
      & 0.4482 & 0.3549 & 0.8724
      \\

      & TruFor
      & 0.6794 & 0.6119 & \underline{0.9618}
      & \underline{0.6980} & \underline{0.6223} & \underline{0.9695}
      & \underline{0.6337} & \underline{0.5585} & \underline{0.9365}
      \\

      & IML-ViT
      & \underline{0.6881} & \underline{0.6246} & 0.8662
      & 0.6555 & 0.5976 & 0.8700
      & 0.6196 & 0.5537 & 0.8720
      \\

      & Mesorch
      & 0.1918 & 0.1529 & 0.8414
      & 0.3171 & 0.2552 & 0.8706
      & 0.2096 & 0.1647 & 0.7764
      \\

      & SparseViT
      & \textbf{0.7847} & \textbf{0.7178} & \textbf{0.9827}
      & \textbf{0.7871} & \textbf{0.7131} & \textbf{0.9866}
      & \textbf{0.7308} & \textbf{0.6582} & \textbf{0.9664}
      \\

    \midrule

    \multirow{5}{*}{\textbf{Replace}}

      & PSCC-Net
      & 0.5196 & 0.4316 & 0.9252
      & 0.4442 & 0.3594 & 0.8724
      & 0.6490 & 0.5644 & 0.9466
      \\

      & TruFor
      & \underline{0.7930} & \underline{0.7234} & \underline{0.9852}
      & \textbf{0.5923} & \underline{0.5190} & \underline{0.9273}
      & \underline{0.8609} & \textbf{0.8010} & \textbf{0.9915}
      \\

      & IML-ViT
      & 0.6933 & 0.6303 & 0.8700
      & 0.4453 & 0.3867 & 0.8096
      & 0.7811 & 0.7282 & 0.9207
      \\

      & Mesorch
      & 0.6656 & 0.6052 & 0.9568
      & 0.4854 & 0.4227 & 0.8854
      & 0.8484 & 0.7913 & 0.9861
      \\

      & SparseViT
      & \textbf{0.8115} & \textbf{0.7485} & \textbf{0.9879}
      & \underline{0.5909} & \textbf{0.5216} & \textbf{0.9313}
      & \textbf{0.8618} & \underline{0.8022} & \underline{0.9909}
      \\

    \bottomrule

  \end{tabular*}
\caption{Complete Protocol~III cross-operation evaluation results (F1/IoU/AUC). Each block groups models by training operation and evaluates all three test operations.}
  \label{tab:intra-domain-p3}
\end{table}

\subsection{Controlled composition/reference results}

Protocol~IV compares single-operation samples with the corresponding Fusion and Multi-Source settings. Table~\ref{tab:benchmark_composition} reports the existing reference results for models trained only on GIFTBench-Core; each composition score is paired with its single-operation reference rather than with an independently sampled test distribution.

\begin{table}[t]
\caption{Protocol~IV reference results for models trained only on GIFTBench-Core. Each composition score is paired with its corresponding single-operation reference. Entries are F1/IoU; $\Delta_{\mathrm{F-S}}$ and $\Delta_{\mathrm{M-S}}$ are the Fusion-minus-Single and Multi-Source-minus-Single differences, respectively.}
\label{tab:benchmark_composition}
\centering
\tiny
\setlength{\tabcolsep}{2.5pt}
\renewcommand{\arraystretch}{0.90}
\begin{tabular}{lcccccc}
\toprule
Model & Single$_{\mathrm{Fusion}}$ & Fusion & Single$_{\mathrm{Multi}}$ & Multi-Source & $\Delta_{\mathrm{F-S}}$ & $\Delta_{\mathrm{M-S}}$ \\
\midrule
PSCC-Net & .611/.511 & .538/.442 & .577/.480 & .486/.394 & -.073/-.069 & -.091/-.086 \\
TruFor & .884/.835 & .814/.729 & .817/.755 & .670/.609 & -.070/-.106 & -.147/-.146 \\
IML-ViT & .673/.649 & .657/.613 & .684/.634 & .606/.558 & -.016/-.036 & -.078/-.076 \\
Mesorch & .794/.719 & .733/.631 & .740/.657 & .707/.613 & -.061/-.088 & -.033/-.044 \\
SparseViT & .908/.862 & .793/.713 & .845/.781 & .722/.653 & -.115/-.149 & -.123/-.128 \\
\bottomrule
\end{tabular}
\end{table}

The comparison shows the effect of composition relative to a matched single-operation reference. It is a reference evaluation for the benchmark protocols, not a ForenScope ablation and not an external-dataset transfer result.

\subsection{Robustness under image degradation}
\label{app:robustness}

\begin{figure}[ht]
    \centering
    \begin{minipage}[t]{0.44\textwidth}
        \centering
        \includegraphics[width=\linewidth]{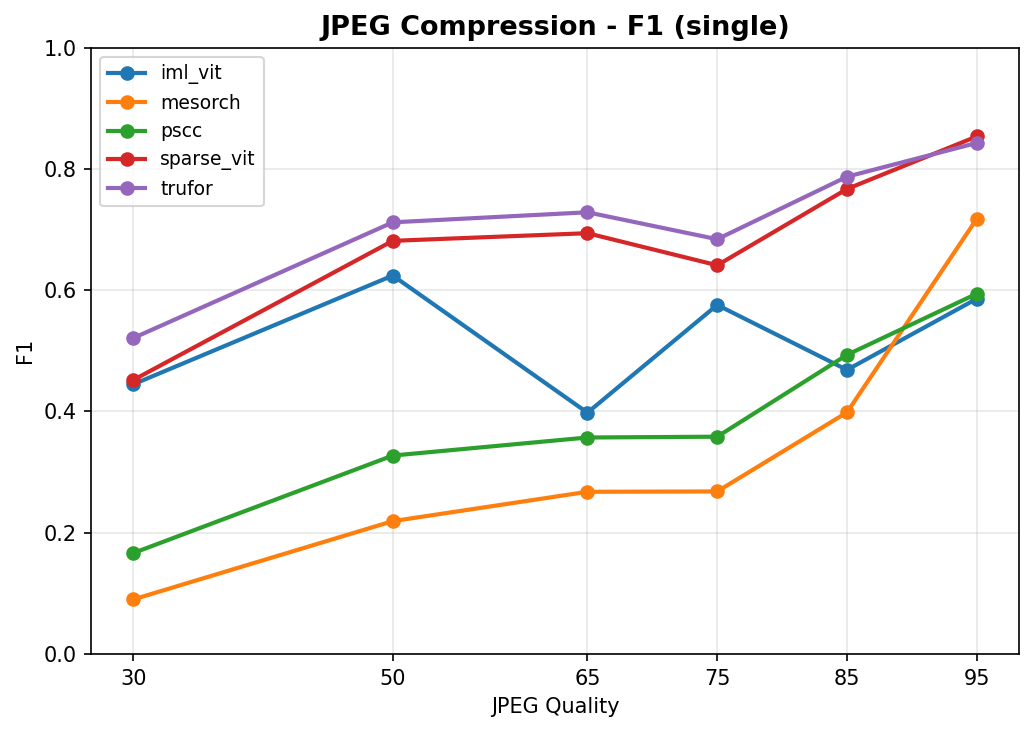}
        \par
        {\footnotesize (a) JPEG compression\par}
    \end{minipage}\hfill
    \begin{minipage}[t]{0.44\textwidth}
        \centering
        \includegraphics[width=\linewidth]{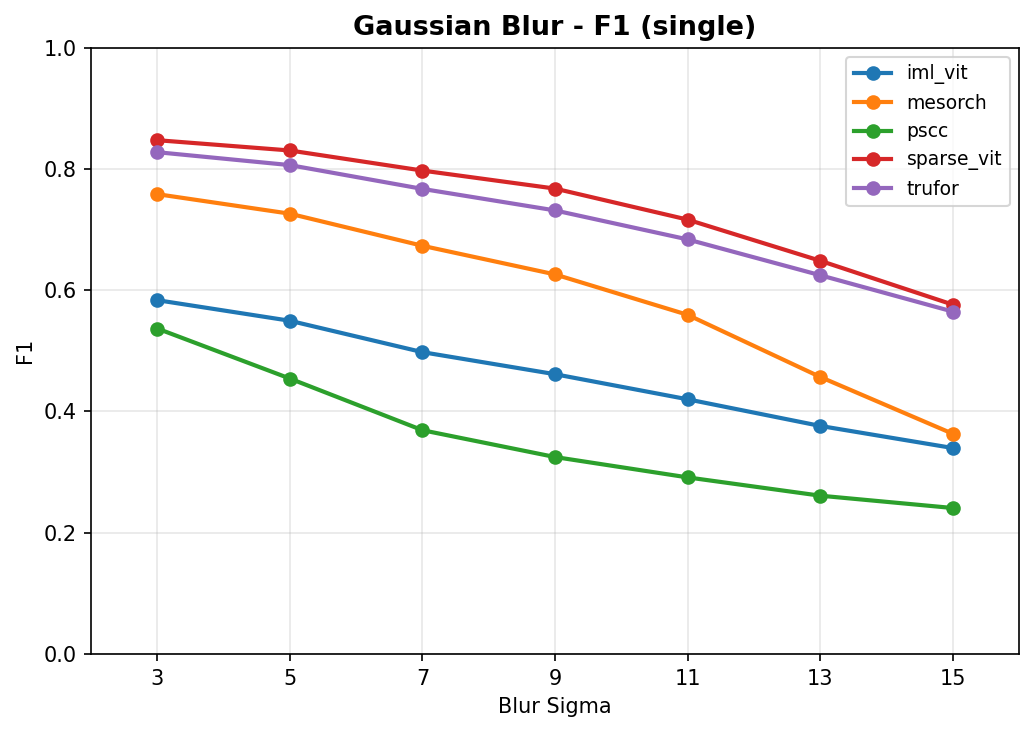}
        \par
        {\footnotesize (b) Gaussian blur\par}
    \end{minipage}
    \caption{Post-processing robustness on single-operation samples under JPEG compression and Gaussian blur.}
    \label{fig:robustness}
\end{figure}

The existing benchmark post-processing stress tests evaluate single-operation images after JPEG compression, Gaussian blur, resize--JPEG, and double-JPEG processing. The JPEG and blur curves in Fig.~\ref{fig:robustness} provide benchmark-level stress evidence; they are not new exp021 Full-Late measurements and are not folded into the four controlled protocol scores.

\begin{table}[ht]
\centering
\scriptsize
\setlength{\tabcolsep}{2.5pt}
\renewcommand{\arraystretch}{0.90}
\caption{Post-processing conditions used in the existing benchmark stress tests. The table defines the perturbations; numerical curves for JPEG and Gaussian blur are shown in Fig.~\ref{fig:robustness}.}
\label{tab:robustness-settings}
\begin{tabular}{ll}
\toprule
Condition & Perturbation \\
\midrule
JPEG & Re-encode the image at the evaluated JPEG quality levels \\
Gaussian blur & Apply the evaluated blur-strength levels to the image \\
Resize--JPEG & Resize before JPEG re-encoding \\
Double-JPEG & Apply two successive JPEG compression stages \\
\bottomrule
\end{tabular}
\end{table}

\section{ForenScope Architecture, Training, and Inference Details}
\label{app:details}

\subsection{Classification adaptation with LoRA}
\label{app:cls-adaptation}

Before localization training, DINOv2 is adapted together with a lightweight image-level classification head. The adaptation stage starts from the official DINOv2 ViT-L/14-reg pretrained weights and inserts LoRA adapters into the attention projections of blocks 8--23. The original encoder weights remain frozen; only the LoRA parameters and the CLS head are optimized with a binary classification loss. The head consumes the CLS token and predicts the real/manipulated image label. The LoRA adapters use a low-rank residual parameterization with a small update scale and dropout, while the classification head uses a larger learning rate than the adapters. Classification checkpoints are selected using the available classification validation protocol.

We use two data configurations for this stage. The standard ForenScope configuration performs the adaptation on the ADE-domain classification data and provides the ADE-adapted encoder and CLS head used by the main localization model. The controlled Protocol-MVSS configuration repeats the same adaptation recipe from the official pretrained encoder on CASIAv2, using its real and manipulated images, and then initializes the subsequent localization stage from the resulting CASIAv2-adapted checkpoint. CASIAv1 is not used to optimize the CASIAv2 classification adaptation. These configurations change the source of the adapted representation, while the downstream localization architecture and the freezing rule remain the same.

After adaptation, the encoder, its LoRA parameters, and the CLS head are frozen during localization training. The localization decoder consumes patch features from the adapted encoder, while the frozen CLS representation is retained for image-level detection and for CLS-conditioned modulation of the localization branch. Thus, the classification adaptation is part of the initialization pipeline rather than an additional loss optimized during localization.

\subsection{Initialization and trainable modules}
\label{app:trainable-modules}

ForenScope initializes its encoder and image-level branch from the ADE-domain classification-adapted DINOv2 model. During localization training, this encoder and the inherited classification branch remain frozen; only the localization-specific modules are optimized. The relevant structural variants and inference settings are summarized below without tying the paper to experiment-directory names or checkpoint hashes.

\begin{table}[H]
\centering
\scriptsize
\setlength{\tabcolsep}{2pt}
\begin{tabular}{p{0.16\textwidth}p{0.24\textwidth}p{0.34\textwidth}p{0.09\textwidth}p{0.12\textwidth}}
\toprule
\textbf{Variant} & \textbf{Encoder / image-level branch} & \textbf{Trainable localization modules} & \textbf{FiLM} & \textbf{Prototype} \\
\midrule
ForenScope & ADE-adapted DINOv2; frozen image-level branch & Multi-depth projections, LayerMixers, decoder, auxiliary/output heads & On & Disabled in default direct output \\
ForenScope + Proto. & Same trained network; frozen image-level branch & Same trained localization branch & On & Enabled at inference only \\
Existing No-CLS & ADE-adapted encoder; frozen image-level branch & Retrained multi-depth branch without CLS-FiLM & Off & Disabled \\
H37-only & ADE-adapted encoder; frozen image-level branch & Retrained H37-only localization variant & Off in effective path & Disabled \\
w/o CLS-FiLM \& ADE adaptation & Original DINOv2; no ADE adaptation & Retrained localization branch without CLS-FiLM & Off & Disabled \\
\bottomrule
\end{tabular}
\caption{Initialization, trainable modules, and inference variants. ForenScope is the default direct output; ForenScope + Proto. uses the same trained network and adds only inference-time prototype refinement. Structural rows are independently retrained unless stated otherwise.}
\label{tab:trainable-modules}
\end{table}

The no-CLS/no-ADE-adaptation row changes both encoder initialization and the CLS-conditioned branch, so it is a joint control rather than a pure classification ablation.

\subsection{Encoder and token extraction}

The encoder is DINOv2 ViT-L/14 with four register tokens. At input resolution \(1036\times1036\), the token sequence is ordered as
\[
\texttt{CLS},\;\texttt{register}_1,\ldots,\texttt{register}_4,\;\texttt{patch}_1,\ldots,\texttt{patch}_N.
\]
The classifier consumes the CLS token. The localization branch removes the first \(1+4\) tokens and reshapes only the remaining patch tokens into spatial maps. Feature taps are the zero-based DINOv2 block indices \(F8,F11,F14,F17\); register tokens are never treated as spatial patches.

\subsection{Multi-scale projection and learnable fusion}

Each of the four tapped feature maps is projected from 1024 to 128 channels by an independent \(1\times1\) projection followed by GroupNorm and GELU. The four projected features are mapped to the three localization resolutions H19, H37, and H74, producing twelve layer--scale lateral paths. Scale-specific downsampling or upsampling preserves the spatial correspondence of each path before fusion.

For each scale, the LayerMixer maintains four learnable logits and normalizes them with a softmax. The resulting output is a four-way convex mixture \(F_s=\sum_l\bar{\alpha}_{l,s}U_{l,s}\), with one learned coefficient vector for each spatial scale. These coefficients are shared across images at inference and are optimized jointly with the localization branch. The fixed-equal intervention in Appendix~\ref{app:ablation-details} replaces these learned coefficients by \(0.25\) for each depth at inference without retraining.

\subsection{CLS-conditioned localization and decoder}

The CLS token is mapped by a two-layer MLP to a conditioning vector. At H19 and H37, FiLM uses a bounded residual modulation
\[
\widetilde{F}_s=F_s+\rho_s\bigl(g_s\odot F_s+r_s\bigr),\qquad
g_s=\tanh(W_{\gamma,s}c+b_{\gamma,s}),\quad
r_s=\tanh(W_{\beta,s}c+b_{\beta,s}),
\]
with $\rho_{19}=1$, $\rho_{37}=0.25$, and no FiLM at H74. The FiLM projections are zero-initialized, so the conditioned path starts as a residual perturbation of the unconditioned representation. This placement leaves the highest-resolution H74 path available for fine spatial evidence.

The decoder fuses the conditioned H19 and H37 features from coarse to fine, then combines the result with the unconditioned H74 feature. Progressive upsampling produces intermediate resolutions 148, 296, 592, and the final 1036 image resolution. Auxiliary heads supervise H19, H37, and H74; the final one-channel output head produces the direct localization logit map.

\subsection{Objective, sampling, augmentation, optimization, and selection}

The localization objective is binary cross-entropy plus soft Dice at the final output and auxiliary scales:
\[
\mathcal{L}=\ell_{\mathrm{final}}+0.1\ell_{19}+0.2\ell_{37}+0.3\ell_{74}.
\]
Manipulated images have weight 1.0 and real images have weight 0.3. The main training protocol samples 17,000 examples from each of Splicing, Copy--Move, Removal, and AI-Edit, includes 532 attribute examples, and samples 20,000 real images, for a planned 88,532 samples per epoch (88,512 effective samples under the three-process DDP run). Validation contains 493 manipulated and 493 real images.

The formal configuration uses three-process DDP with batch size 8 per process (global batch size 24), two data-loader workers per process, five epochs, learning rate \(3\times10^{-4}\), and seed 42. Only the 3,210,256 localization-branch parameters are optimized. We use AdamW with weight decay 0.05, clip the decoder gradient norm at 1.0, linearly warm up for 500 optimizer steps, and then apply cosine decay to a learning-rate floor of \(5\times10^{-7}\). EMA with decay 0.999 is updated after every optimizer step. Geometry-changing augmentation is applied jointly to image and mask; photometric augmentation is image-only; masks use nearest-neighbor resizing and are re-binarized. The best checkpoint is selected from held-out in-domain pixel F1 only, with no OOD labels used for checkpoint selection. Both live and EMA weights are saved, but the reported standard localization tables use the selected EMA weights unless a diagnostic row states otherwise.

All twelve external datasets are excluded from optimization, validation, and checkpoint selection; they are used only by the evaluation scripts after model selection.

\subsection{Prototype-guided inference}
\label{app:prototype}

Prototype refinement is an optional inference-only strategy evaluated in addition to the default direct ForenScope output. It does not update the encoder, decoder, optimizer state, or checkpoint. For each test image, the procedure is:

\begin{enumerate}
\item obtain the direct localization logits and the normalized H37 patch features;
\item use the direct H37 prediction to form manipulated and real support features with the fixed 70/30 trimmed support rule;
\item compute the cosine-similarity margin between each H37 patch and the two support prototypes;
\item if a support set is empty, use the direct H37 probability \(q_i\) for the manipulated support or \(1-q_i\) for the real support as its soft fallback; add the calibrated prototype correction with the fixed \(+8\) logit scale, and upsample to the output mask;
\item threshold the refined mask at 0.5.
\end{enumerate}

More formally, let \(h_i\) denote the \(\ell_2\)-normalized H37 feature at spatial position \(i\), and let \(q_i\in[0,1]\) denote the direct localization probability resized to the H37 grid. For the reported \texttt{final\_trim\_70\_30\_raw} configuration, confident manipulated and real regions are assigned the weights
\begin{equation}
w_i^+=\max(q_i-0.7,0),
\qquad
w_i^-=\max(0.3-q_i,0).
\end{equation}
The corresponding prototypes are the normalized weighted means
\begin{equation}
\mu^+=\frac{\sum_i w_i^+h_i}{\left\lVert\sum_i w_i^+h_i\right\rVert_2},
\qquad
\mu^-=\frac{\sum_i w_i^-h_i}{\left\lVert\sum_i w_i^-h_i\right\rVert_2}.
\end{equation}
If either confident support set is empty, its weights are replaced by the soft fallback \(q_i\) for the manipulated prototype or \(1-q_i\) for the real prototype. The cosine-similarity margin at each H37 position is
\begin{equation}
m_i=\langle h_i,\mu^+\rangle-\langle h_i,\mu^-\rangle.
\end{equation}
If \(L\) denotes the direct full-resolution localization logit, the refined logit is
\begin{equation}
L^{\mathrm{ref}}=L+8\,\operatorname{Up}_{37\rightarrow H}(m),
\end{equation}
where \(\operatorname{Up}_{37\rightarrow H}\) denotes bilinear upsampling to the output resolution. The refined mask is obtained by thresholding \(\sigma(L^{\mathrm{ref}})\) at 0.5.

The fixed \texttt{final\_trim\_70\_30\_raw} configuration is applied unchanged across the twelve external datasets. Prototype computation uses the model's own predictions and features, not target-domain labels, and the refinement does not update any model parameters. ``ForenScope'' denotes the direct network output in all primary tables, whereas ``ForenScope + Proto.'' denotes the same checkpoint followed by this fixed inference correction.

\section{Complete ForenScope Evaluation and Ablations}
\label{app:forenscope-results}

This section gathers the ForenScope evaluation that is summarized in the main paper. The default method is the learned localization network and its direct detection and localization outputs. An optional fixed prototype-guided inference strategy is reported separately as an auxiliary variant. Results distinguish in-domain localization, external localization, image-level detection, retrained structural variants, and fixed-checkpoint interventions.

\subsection{Evaluation protocol for ForenScope}

In-domain evaluation uses \texttt{cross/in\_domain/test\_all.json}, whose 13 masked groups are organized into source (Splicing, Copy--Move, Removal, AI-Edit), semantic target (Object, Component, Background), operation (Add, Remove, Replace), and composition (Single, Fusion, Multi-Source) views. External localization uses the primary-6 and extra-6 groups listed in Appendix~\ref{app:metric-aggregation}; All-12 is their equal-weight macro average. The reported ForenScope row is the direct output, while ForenScope + Proto. uses the same checkpoint with the optional inference correction.

\subsection{Complete external localization}

Table~\ref{tab:ood_full} reports all twelve external datasets and the optional inference comparison. The ForenScope row is the direct network output, whereas ForenScope + Proto. applies the same fixed checkpoint followed by prototype refinement. The table therefore preserves the distinction between learned-network output and inference-time refinement rather than hiding the per-dataset trade-off.

\begin{table}[t]
\centering
\scriptsize
\setlength{\tabcolsep}{2.4pt}
\renewcommand{\arraystretch}{0.9}
\caption{External localization results with methods as rows. Entries are F1/IoU; ForenScope is the direct output and ForenScope + Proto. adds fixed inference-only refinement. Primary-6, Extra-6, and All-12 are macro averages over the corresponding columns.}
\label{tab:ood_full}
\resizebox{\textwidth}{!}{%
\begin{tabular}{l*{15}{c}}
\toprule
& \multicolumn{6}{c}{Primary-6} & \multicolumn{6}{c}{Extra-6} & \multicolumn{3}{c}{Summary} \\
\cmidrule(lr){2-7}\cmidrule(lr){8-13}\cmidrule(lr){14-16}
Method & CASIA & CocoGlide & Columbia & IMD2020 & NIST16 & COVERAGE & MISD & AutoSplice & BRGen & SceneText & OSTF & DDL & Primary-6 & Extra-6 & All-12 \\
\midrule
ForenScope & .679/.598 & .799/.697 & .546/.452 & .579/.488 & .548/.469 & .578/.472 & .696/.560 & .871/.789 & .746/.665 & .703/.596 & .595/.476 & .520/.412 & .621/.529 & .688/.583 & .655/.556 \\
Existing No-CLS & .660/.582 & .762/.657 & .534/.442 & .572/.482 & .516/.438 & .581/.467 & .694/.562 & .897/.827 & .759/.679 & .724/.617 & .615/.497 & .486/.386 & .604/.511 & .696/.595 & .650/.553 \\
H37-only & .658/.577 & .676/.564 & .495/.401 & .570/.477 & .547/.461 & .494/.384 & .688/.553 & .889/.816 & .748/.666 & .691/.592 & .578/.465 & .527/.416 & .573/.477 & .687/.584 & .630/.531 \\
Joint control & .567/.490 & .593/.481 & .601/.499 & .550/.456 & .500/.420 & .424/.313 & .686/.550 & .897/.831 & .748/.661 & .685/.580 & .574/.459 & .463/.363 & .539/.443 & .676/.574 & .607/.509 \\
ForenScope + Proto. & .695/.615 & .820/.724 & .564/.472 & .580/.487 & .566/.485 & .592/.487 & .726/.597 & .873/.793 & .753/.671 & .705/.598 & .600/.478 & .532/.425 & .636/.545 & .698/.594 & .667/.569 \\
\bottomrule
\end{tabular}
}%
\end{table}

\subsection{Image-level detection results}

The detector is the frozen classification head inherited from the ADE adaptation stage; localization training does not optimize its loss. Table~\ref{tab:detection} reports the complete paired AUC/F1 comparison on the eight external datasets containing both real and manipulated images, including the controlled CASIAv2-trained setting. AUC is undefined for BRGen, SceneText, OSTF, and DDL because these splits contain only manipulated images. These results document retained detection capability, not a standalone detection leaderboard.

\begin{table}[t]
\centering
\scriptsize
\setlength{\tabcolsep}{2pt}
\caption{Image-level detection AUC and F1 on eight external datasets containing both classes.}
\label{tab:detection}
\resizebox{\textwidth}{!}{%
\begin{tabular}{ll*{9}{cc}}
\toprule
\multirow{2}{*}{Training Protocol} & \multirow{2}{*}{Model}
& \multicolumn{2}{c}{CASIA v1}
& \multicolumn{2}{c}{CocoGlide}
& \multicolumn{2}{c}{Columbia}
& \multicolumn{2}{c}{IMD2020}
& \multicolumn{2}{c}{NIST16}
& \multicolumn{2}{c}{COVERAGE}
& \multicolumn{2}{c}{MISD}
& \multicolumn{2}{c}{AutoSplice}
& \multicolumn{2}{c}{Avg.} \\
\cmidrule(lr){3-4}\cmidrule(lr){5-6}\cmidrule(lr){7-8}\cmidrule(lr){9-10}\cmidrule(lr){11-12}\cmidrule(lr){13-14}\cmidrule(lr){15-16}\cmidrule(lr){17-18}\cmidrule(lr){19-20}
& & AUC & F1 & AUC & F1 & AUC & F1 & AUC & F1 & AUC & F1 & AUC & F1 & AUC & F1 & AUC & F1 & AUC & F1 \\
\midrule
\multirow{4}{*}{\textbf{Protocol-MVSS}}
& MVSS-Net & .639 & .697 & .507 & .667 & .468 & .663 & .564 & .907 & .500 & .943 & .519 & \underline{.667} & .704 & .489 & .543 & .761 & .556 & .724 \\
& PSCC-Net & .743 & .521 & .440 & .639 & .806 & \underline{.732} & .629 & \underline{.898} & .547 & \underline{.887} & .495 & .662 & \textbf{.994} & \underline{.869} & \underline{.827} & \underline{.822} & .685 & \underline{.754} \\
& TruFor & .148 & .392 & .446 & .625 & .049 & .100 & .392 & .796 & .425 & .819 & .352 & .592 & .034 & .075 & .459 & .646 & .288 & .506 \\
& \method & \underline{.875} & \underline{.732} & \underline{.702} & \underline{.675} & \underline{.819} & .731 & \underline{.727} & .815 & \underline{.563} & .688 & \underline{.551} & .418 & .892 & .779 & .691 & .400 & \underline{.727} & .655 \\
\midrule
\textbf{GIFTBench} & \textbf{\method} & \textbf{.902} & \textbf{.844} & \textbf{.989} & \textbf{.963} & \textbf{.994} & \textbf{.967} & \textbf{.899} & .834 & \textbf{.837} & .705 & \textbf{.891} & \textbf{.809} & \underline{.993} & \textbf{.911} & \textbf{.977} & \textbf{.934} & \textbf{.935} & \textbf{.871} \\
\bottomrule
\end{tabular}
}
\end{table}

The same frozen ADE-adapted encoder and classification head are shared by the Full-Late, Existing No-CLS, and H37-only localization variants. Removing CLS-FiLM or changing the localization decoder therefore does not create a new detector, so the detection table is not duplicated for those localization variants.

\subsection{Ablations and mechanism analysis}
\label{app:ablation-details}

The following tables separate independently retrained structural variants from fixed-checkpoint inference interventions. This distinction is essential: the structural table evaluates the final retrained Static LayerMixer model, whereas the diagnostic table changes only the prediction path of separately identified reference checkpoints. Thus, the diagnostic rows are not a second evaluation of the final ForenScope checkpoint. All structural rows use direct network outputs; the final inference row adds the prototype without retraining.

\begin{table}[ht]
\centering
\scriptsize
\setlength{\tabcolsep}{1.8pt}
\renewcommand{\arraystretch}{0.78}
\caption{Definitions of the principal structural variants and inference interventions used in the complete evaluation.}
\label{tab:ablation_def}
\resizebox{0.90\linewidth}{!}{%
\begin{tabular}{p{0.25\linewidth}p{0.63\linewidth}}
\toprule
Variant & Definition \\
\midrule
ForenScope & Full multi-depth network with learned LayerMixer and CLS-FiLM; direct output only. \\
Existing No-CLS & ADE-adapted encoder retained, but the localization branch is retrained without CLS-FiLM. \\
H37-only & H37 is the only effective spatial scale; the remaining fusion topology is retained. \\
w/o ADE adaptation & Original DINOv2 encoder with the full multi-scale, CLS-conditioned localization architecture. \\
Joint control & Original DINOv2 initialization, no ADE adaptation, and no CLS-FiLM; retrained jointly. \\
Equal fixed mixer & Fixed-checkpoint intervention replacing learned four-depth weights by equal weights. \\
FiLM-off / CLS-shuffle & Fixed original Full-Late checkpoint interventions disabling FiLM or shuffling the CLS conditioning. \\
ForenScope + Proto. & Fixed inference-time refinement applied after the direct network output. \\
\bottomrule
\end{tabular}
}%
\end{table}

\begin{table}[H]
\centering
\scriptsize
\setlength{\tabcolsep}{2pt}
\renewcommand{\arraystretch}{0.82}
\caption{Retrained structural ablations. ForenScope denotes the final Static LayerMixer checkpoint; entries are direct-output F1/IoU, with in-domain as the thirteen-group macro average and OOD as All-12.}
\label{tab:ablation-retrained-full}
\begin{tabular}{lcc}
\toprule
Variant & In-domain & OOD \\
\midrule
ForenScope & .886/.823 & .655/.556 \\
Existing No-CLS & .881/.816 & .650/.553 \\
H37-only & .871/.801 & .630/.531 \\
w/o ADE adaptation & -- & .618/.519 \\
Joint control & .873/.805 & .607/.509 \\
\bottomrule
\end{tabular}
\end{table}

The pure no-adaptation variant was evaluated on All-12 but not on the thirteen-group in-domain test, so its in-domain entry is left undefined. The structural rows show that ADE adaptation and multi-scale evidence improve transfer, whereas removing CLS-FiLM has a smaller effect after retraining. The joint control changes both encoder adaptation and conditioning and therefore does not isolate either factor.

\begin{table}[H]
\centering
\scriptsize
\setlength{\tabcolsep}{2pt}
\renewcommand{\arraystretch}{0.82}
\caption{Fixed-checkpoint mechanism diagnostics on All-12. The equal-layer control uses the final Static LayerMixer checkpoint; the No-CLS and Full-Late rows use their respective EMA checkpoints. None of these rows retrains the network.}
\label{tab:ablation-interventions}
\begin{tabular}{llc}
\toprule
Reference checkpoint (EMA) & Inference condition & F1/IoU \\
\midrule
ForenScope (Static) & Equal layer weights & .653/.557 \\
Existing No-CLS & Learned layer fusion & .650/.553 \\
Existing No-CLS & Equal layer weights & .553/.450 \\
Original Full-Late & Normal CLS-FiLM & .643/.548 \\
Original Full-Late & FiLM off & .622/.518 \\
Original Full-Late & CLS shuffled & .633/.536 \\
\bottomrule
\end{tabular}
\end{table}

These fixed-checkpoint interventions test whether the corresponding trained prediction paths use learned fusion and CLS conditioning, rather than comparing independently retrained architectures. In particular, the \emph{Original Full-Late} normal row (.643/.548) is a separate diagnostic reference and should not be compared as the Full baseline of Table~\ref{tab:ablation-retrained-full}, whose final Static LayerMixer ForenScope row is .655/.556. A matched FiLM intervention on the final Static LayerMixer checkpoint would require a separate inference run. The optional prototype refinement is parameter-free and improves the external All-12 average, but its in-domain evaluation shows that the gain is not uniform across benchmark groups.

\subsection{Reproducibility checklist}

The released experiment records specify: DINOv2 ViT-L/14 reg4; input resolution \(1036\times1036\); feature taps \(F8/F11/F14/F17\); H19/H37/H74 spatial scales; threshold 0.5; EMA decay 0.999; seed 42; live/EMA checkpoint distinction; the fixed prototype rule; the single-seed policy; no target-domain fine-tuning; in-domain-only checkpoint selection; and the in-domain and external evaluation manifests. Prototype inference uses the model's own predictions and features with shared fixed hyperparameters, and requires neither target-domain labels nor parameter updates.

\section{Qualitative Analysis}
\label{app:qualitative}

This section complements the numerical results with visual evidence of transfer behavior and failure cases.

\subsection{Dataset-level transfer and representation views}

\begin{figure}[H]
    \centering
    \includegraphics[width=0.72\textwidth]{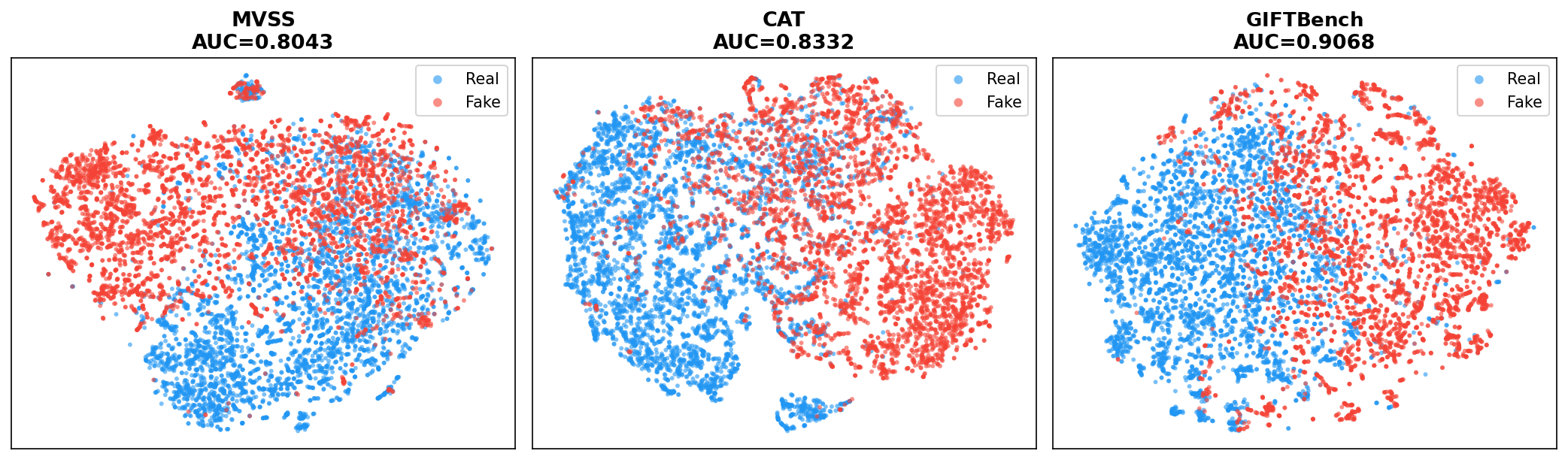}
    \par
    {\footnotesize (a) t-SNE view of representative decoder features\par}
    \caption{Supplementary representation visualization of TruFor decoder features under different benchmark training conditions.}
    \label{fig:app-external-analysis}
\end{figure}

Figure~\ref{fig:app-external-analysis} provides a descriptive two-dimensional projection of TruFor decoder features under different training distributions.

\subsection{External qualitative comparisons}
\label{app:benchmark-external-details}

\begin{figure}[H]
    \centering
    \includegraphics[width=0.90\columnwidth]{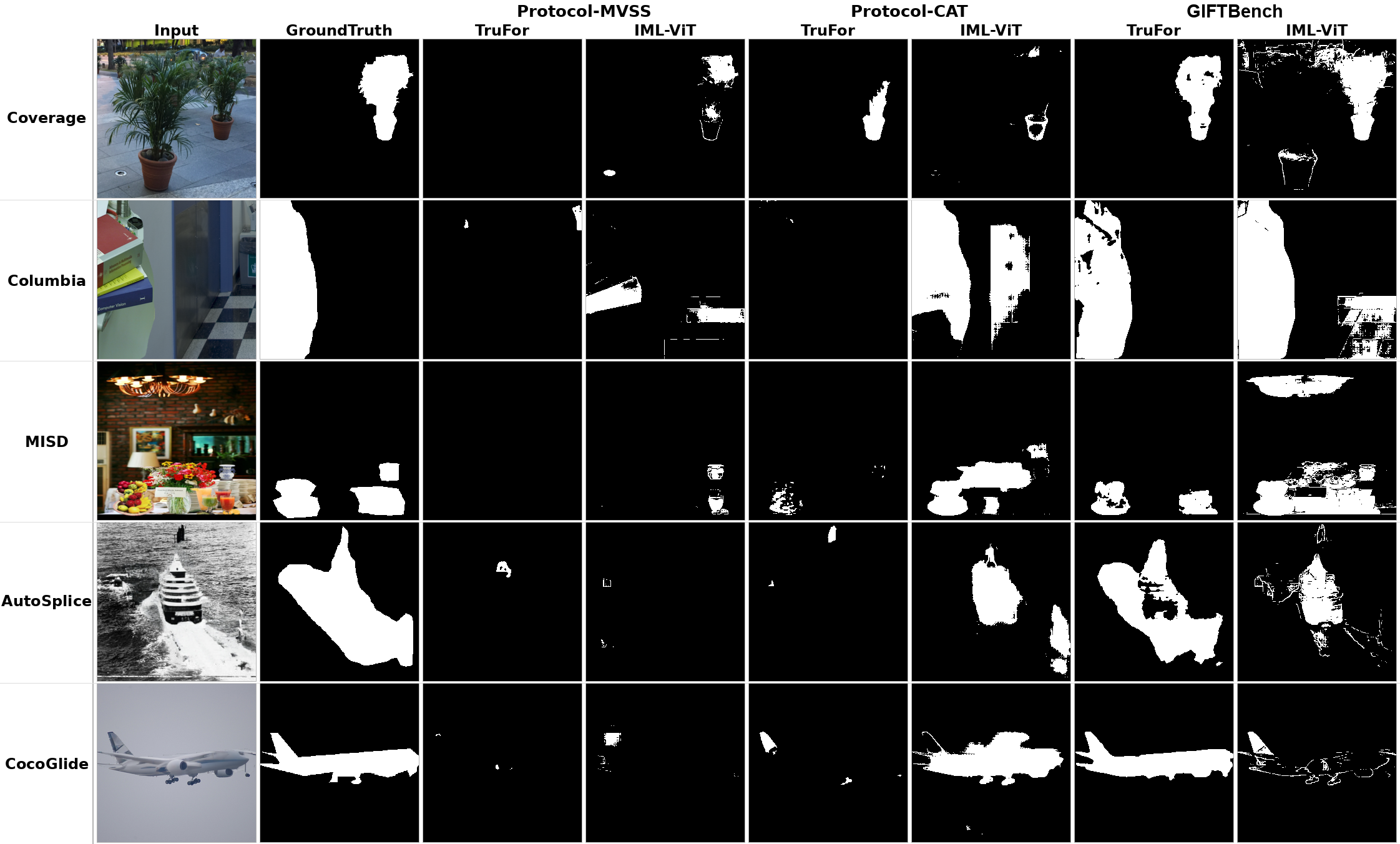}
    \caption{Qualitative external-transfer comparison under Protocol-MVSS, Protocol-CAT, and GIFTBench training.}
    \label{fig:app-external-qualitative}
\end{figure}

Figure~\ref{fig:app-external-qualitative} qualitatively compares representative models trained under Protocol-MVSS, Protocol-CAT, and GIFTBench; these examples are not used for model selection.

\subsection{Good and bad cases on GIFTBench}

\begin{figure}[H]
    \centering
    \begin{minipage}[t]{0.43\textwidth}
        \centering
        \includegraphics[width=\linewidth]{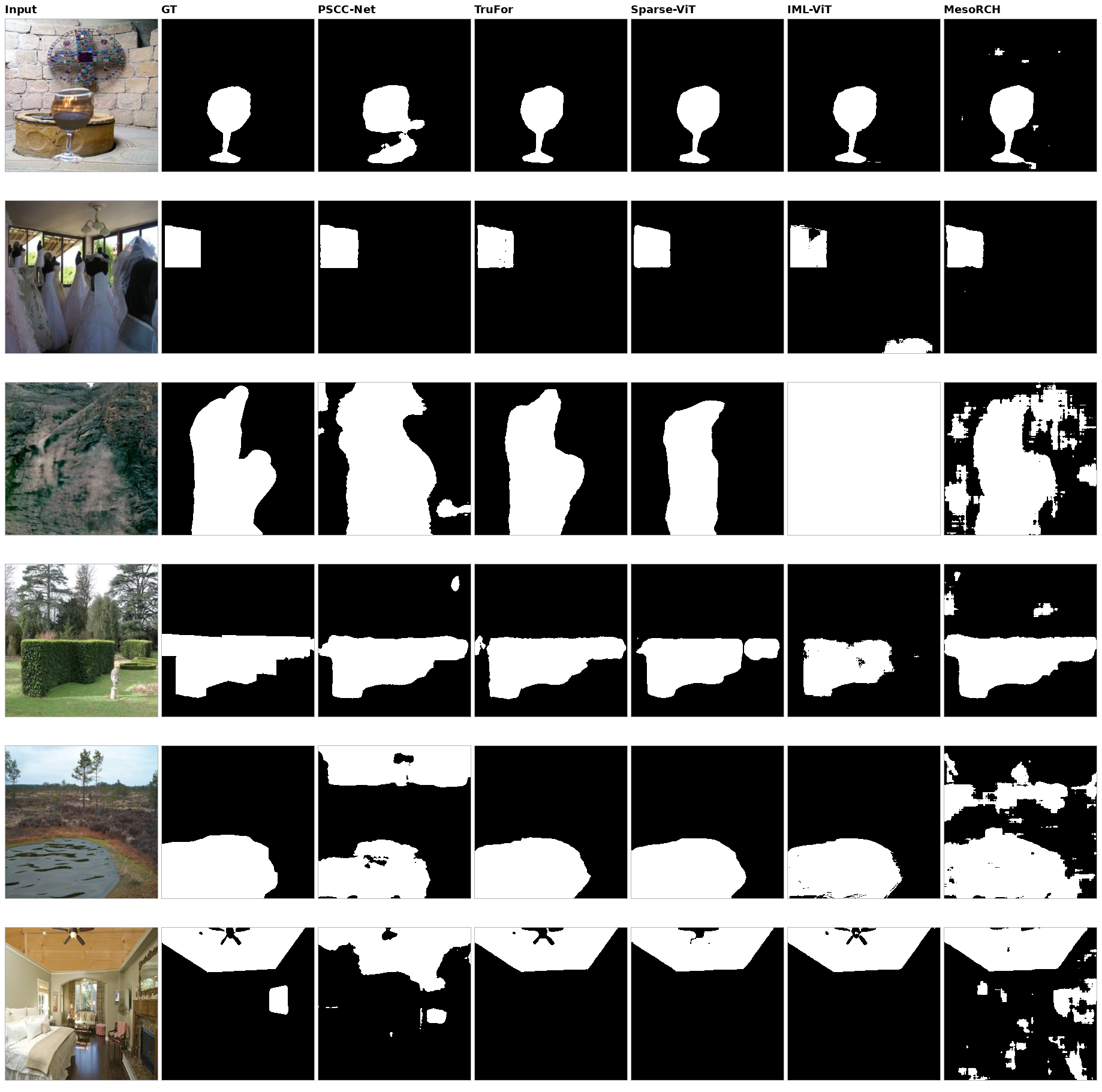}
        \par
        {\footnotesize (a) Representative high-agreement cases\par}
    \end{minipage}\hfill
    \begin{minipage}[t]{0.43\textwidth}
        \centering
        \includegraphics[width=\linewidth]{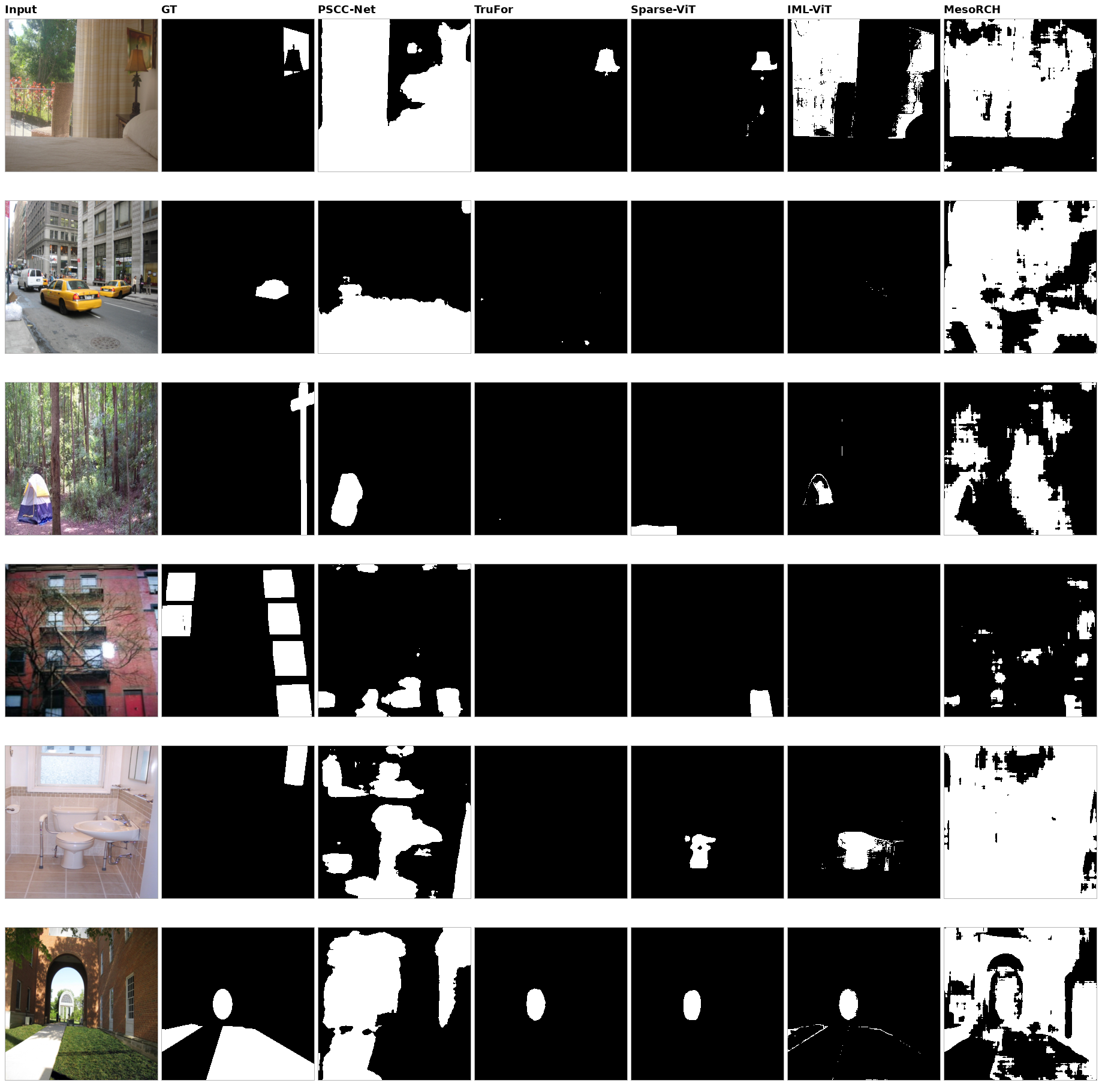}
        \par
        {\footnotesize (b) Representative all-model failure cases\par}
    \end{minipage}
    \caption{Representative high-agreement and all-model failure cases on GIFTBench.}
    \label{fig:benchmark-good-bad}
\end{figure}

Figure~\ref{fig:benchmark-good-bad} contrasts high-agreement benchmark examples with cases for which all displayed baseline models obtain F1 below 0.15.

\clearpage
\bibliographystyle{plainnat}
\bibliography{references,imdl_refs}

@article{oquab2024dinov2,
  title={{DINOv2}: Learning Robust Visual Features without Supervision},
  author={Oquab, Maxime and Darcet, Timoth{\'e}e and Moutakanni, Th{\'e}o and Vo, Huy V. and Szafraniec, Marc and Khalidov, Vasil and Fernandez, Pierre and Haziza, Daniel and Massa, Francisco and El-Nouby, Alaa and Assran, Mahmoud and Ballas, Nicolas and Galuba, Wojciech and Howes, Russell and Huang, Po-Yao and Li, Shang-Wen and Misra, Ishan and Rabbat, Michael and Sharma, Vasu and Synnaeve, Gabriel and Xu, Hu and J{\'e}gou, Herv{\'e} and Mairal, Julien and Labatut, Patrick and Joulin, Armand and Bojanowski, Piotr},
  journal={Transactions on Machine Learning Research},
  year={2024}
}

@inproceedings{dong2013casia,
  title={Casia image tampering detection evaluation database},
  author={Dong, Jing and others},
  booktitle={IEEE China summit and international conference on signal and information processing},
  pages={422--426},
  year={2013},
}

@inproceedings{hsu2006detecting,
  title={Detecting image splicing using geometry invariants and camera characteristics consistency},
  author={Hsu, Yu-Feng and others},
  booktitle={IEEE international conference on multimedia and expo},
  pages={549--552},
  year={2006},
}

@inproceedings{wen2016coverage,
  title={COVERAGE---A novel database for copy-move forgery detection},
  author={Wen, Bihan and Zhu, Ye and Subramanian, Ramanathan and Ng, Tian-Tsong and Shen, Xuanjing and Winkler, Stefan},
  booktitle={IEEE international conference on image processing},
  pages={161--165},
  year={2016},
}

@inproceedings{guan2019mfc,
  title={MFC datasets: Large-scale benchmark datasets for media forensic challenge evaluation},
  author={Guan, Haiying and Kozak, Mark and Robertson, Eric and Lee, Yooyoung and Yates, Amy N and Delgado, Andrew and Zhou, Daniel and Kheyrkhah, Timothee and Smith, Jeff and Fiscus, Jonathan},
  booktitle={IEEE Winter Applications of Computer Vision Workshops},
  pages={63--72},
  year={2019},
  organization={IEEE}
}

@inproceedings{novozamsky2020imd2020,
  title={IMD2020: A large-scale annotated dataset tailored for detecting manipulated images},
  author={Novozamsky, Adam and others},
  booktitle={Proceedings of the IEEE/CVF winter conference on applications of computer vision workshops},
  pages={71--80},
  year={2020}
}

@inproceedings{tralic2013comofod,
  title={CoMoFoD---New database for copy-move forgery detection},
  author={Tralic, Dijana and Zupancic, Ivan and Grgic, Sonja and Grgic, Mislav},
  booktitle={Proceedings ELMAR-2013},
  pages={49--54},
  year={2013},
}

@inproceedings{mahfoudi2019defacto,
  title={Defacto: Image and face manipulation dataset},
  author={Mahfoudi, Ga{\"e}l and Tajini, Badr and Retraint, Florent and Morain-Nicolier, Frederic and Dugelay, Jean Luc and Pic, Marc},
  booktitle={27Th european signal processing conference},
  pages={1--5},
  year={2019},
}

@inproceedings{mareen2024tgif,
  title={TGIF: Text-guided inpainting forgery dataset},
  author={Mareen, Hannes and Karageorgiou, Dimitrios and Van Wallendael, Glenn and Lambert, Peter and Papadopoulos, Symeon},
  booktitle={IEEE International Workshop on Information Forensics and Security},
  pages={1--6},
  year={2024},
}

@inproceedings{jia2023autosplice,
  title={Autosplice: A text-prompt manipulated image dataset for media forensics},
  author={Jia, Shan and Huang, Mingzhen and Zhou, Zhou and Ju, Yan and Cai, Jialing and Lyu, Siwei},
  booktitle={Proceedings of the IEEE/CVF conference on computer vision and pattern recognition},
  pages={893--903},
  year={2023}
}

@misc{xu2025manipshield,
  author       = {Xu, Zitong and Duan, Huiyu and Wang, Xiaoyu and Cai, Zhaolin and Zhang, Kaiwei and Hu, Qiang and Liu, Jing and Min, Xiongkuo and Zhai, Guangtao},
  title        = {ManipShield: A Unified Framework for Image Manipulation Detection, Localization and Explanation},
  year         = {2025},
  eprint       = {2511.14259},
  archivePrefix = {arXiv},
  primaryClass = {cs.CV},
}

@inproceedings{chen2025gim,
  title={Gim: A million-scale benchmark for generative image manipulation detection and localization},
  author={Chen, Yirui and Huang, Xudong and Zhang, Quan and Li, Wei and Zhu, Mingjian and Yan, Qiangyu and Li, Simiao and Chen, Hanting and Hu, Hailin and Yang, Jie and others},
  booktitle={Proceedings of the AAAI Conference on Artificial Intelligence},
  volume={39},
  number={2},
  pages={2311--2319},
  year={2025}
}

@inproceedings{cai2026zooming,
  title={Zooming in on fakes: A novel dataset for localized ai-generated image detection with forgery amplification approach},
  author={Cai, Lvpan and Wang, Haowei and Ji, Jiayi and Zhoumen, Yanshu and Chen, Shen and Yao, Taiping and Sun, Xiaoshuai},
  booktitle={Proceedings of the AAAI Conference on Artificial Intelligence},
  volume={40},
  number={4},
  pages={2534--2542},
  year={2026}
}

@inproceedings{guillaro2023trufor,
  title={Trufor: Leveraging all-round clues for trustworthy image forgery detection and localization},
  author={Guillaro, Fabrizio and Cozzolino, Davide and Sud, Avneesh and Dufour, Nicholas and Verdoliva, Luisa},
  booktitle={Proceedings of the IEEE/CVF conference on computer vision and pattern recognition},
  pages={20606--20615},
  year={2023}
}

@inproceedings{hu2020span,
  title={SPAN: Spatial pyramid attention network for image manipulation localization},
  author={Hu, Xuefeng and Zhang, Zhihan and Jiang, Zhenye and Chaudhuri, Syomantak and Yang, Zhenheng and Nevatia, Ram},
  booktitle={European conference on computer vision},
  pages={312--328},
  year={2020},
}

@inproceedings{su2025can,
  title={Can we get rid of handcrafted feature extractors? sparsevit: Nonsemantics-centered, parameter-efficient image manipulation localization through spare-coding transformer},
  author={Su, Lei and Ma, Xiaochen and Zhu, Xuekang and Niu, Chaoqun and Lei, Zeyu and Zhou, Ji-Zhe},
  booktitle={Proceedings of the AAAI conference on artificial intelligence},
  volume={39},
  number={7},
  pages={7024--7032},
  year={2025}
}

@misc{ma2023iml,
  author       = {Ma, Xiaochen and Du, Bo and Jiang, Zhuohang and Hammadi, Ahmed Y Al and Zhou, Jizhe},
  title        = {IML-ViT: Benchmarking image manipulation localization by vision transformer},
  year         = {2023},
  eprint       = {2307.14863},
  archivePrefix = {arXiv},
  primaryClass = {cs.CV},
}

@inproceedings{zhu2025mesoscopic,
  title={Mesoscopic insights: orchestrating multi-scale \& hybrid architecture for image manipulation localization},
  author={Zhu, Xuekang and Ma, Xiaochen and Su, Lei and Jiang, Zhuohang and Du, Bo and Wang, Xiwen and Lei, Zeyu and Feng, Wentao and Pun, Chi-Man and Zhou, Ji-Zhe},
  booktitle={Proceedings of the AAAI conference on artificial intelligence},
  volume={39},
  number={10},
  pages={11022--11030},
  year={2025}
}

@inproceedings{wu2019mantra,
  title={Mantra-net: Manipulation tracing network for detection and localization of image forgeries with anomalous features},
  author={Wu, Yue and others},
  booktitle={Proceedings of the IEEE/CVF conference on computer vision and pattern recognition},
  pages={9543--9552},
  year={2019}
}

@inproceedings{chen2021image,
  title={Image manipulation detection by multi-view multi-scale supervision},
  author={Chen, Xinru and Dong, Chengbo and Ji, Jiaqi and Cao, Juan and Li, Xirong},
  booktitle={Proceedings of the IEEE/CVF international conference on computer vision},
  pages={14185--14193},
  year={2021}
}

@inproceedings{wang2022objectformer,
  title={Objectformer for image manipulation detection and localization},
  author={Wang, Junke and Wu, Zuxuan and Chen, Jingjing and Han, Xintong and Shrivastava, Abhinav and Lim, Ser-Nam and Jiang, Yu-Gang},
  booktitle={Proceedings of the IEEE/CVF conference on computer vision and pattern recognition},
  pages={2364--2373},
  year={2022}
}

@inproceedings{xu2024fakeshield,
  title={Fakeshield: Explainable image forgery detection and localization via multi-modal large language models},
  author={Xu, Zhipei and Zhang, Xuanyu and Li, Runyi and Tang, Zecheng and Huang, Qing and Zhang, Jian},
  booktitle={International Conference on Learning Representations},
  volume={2025},
  pages={31186--31216},
  year={2025}
}

@article{bayar2018constrained,
  title={Constrained convolutional neural networks: A new approach towards general purpose image manipulation detection},
  author={Bayar, Belhassen and Stamm, Matthew C},
  journal={IEEE Transactions on Information Forensics and Security},
  volume={13},
  number={11},
  pages={2691--2706},
  year={2018},
}

@inproceedings{kirillov2023segment,
  title={Segment anything},
  author={Kirillov, Alexander and Mintun, Nikhila and Ravi, Nikhila and Mao, Hanzi and Rolland, Chloe and Gustafson, Laura and Xiao, Tete and Whitehead, Spencer and Berg, Alexander C and Lo, Wan-Yen and others},
  booktitle={Proceedings of the IEEE/CVF international conference on computer vision},
  pages={4015--4026},
  year={2023}
}

@inproceedings{ravi2024sam,
  title={Sam 2: Segment anything in images and videos},
  author={Ravi, Nikhila and Gabeur, Valentin and Hu, Yuan-Ting and Hu, Ronghang and Ryali, Chaitanya and Ma, Tengyu and Khedr, Haitham and R{\"a}dle, Roman and Rolland, Chloe and Gustafson, Laura and others},
  booktitle={International Conference on Learning Representations},
  volume={2025},
  pages={28085--28128},
  year={2025}
}

@article{zhou2019semantic,
  title={Semantic understanding of scenes through the ade20k dataset},
  author={Zhou, Bolei and Zhao, Hang and Puig, Xavier and Xiao, Tete and Fidler, Sanja and Barriuso, Adela and Torralba, Antonio},
  journal={International journal of computer vision},
  volume={127},
  number={3},
  pages={302--321},
  year={2019},
}

@misc{bai2025qwen25vltechnicalreport,
      title={Qwen2.5-VL Technical Report}, 
      author={Shuai Bai and Keqin Chen and Xuejing Liu and Jialin Wang and Wenbin Ge and Sibo Song and Kai Dang and Peng Wang and Shijie Wang and Jun Tang and Humen Zhong and Yuanzhi Zhu and Mingkun Yang and Zhaohai Li and Jianqiang Wan and Pengfei Wang and Wei Ding and Zheren Fu and Yiheng Xu and Jiabo Ye and Xi Zhang and Tianbao Xie and Zesen Cheng and Hang Zhang and Zhibo Yang and Haiyang Xu and Junyang Lin},
      year={2025},
      eprint={2502.13923},
      archivePrefix={arXiv},
      primaryClass={cs.CV},
      url={https://arxiv.org/abs/2502.13923}, 
}

@misc{bai2025qwen3,
  author       = {Bai, Shuai and Cai, Yuxuan and Chen, Ruizhe and Chen, Keqin and Chen, Xionghui and Cheng, Zesen and Deng, Lianghao and Ding, Wei and Gao, Chang and Ge, Chunjiang and others},
  title        = {Qwen3-vl technical report},
  year         = {2025},
  eprint       = {2511.21631},
  archivePrefix = {arXiv},
  primaryClass = {cs.CV},
}

@misc{wang2025internvl3,
  author       = {Wang, Weiyun and Gao, Zhangwei and Gu, Lixin and Pu, Hengjun and Cui, Long and Wei, Xingguang and Liu, Zhaoyang and Jing, Linglin and Ye, Shenglong and Shao, Jie and others},
  title        = {Internvl3. 5: Advancing open-source multimodal models in versatility, reasoning, and efficiency},
  year         = {2025},
  eprint       = {2508.18265},
  archivePrefix = {arXiv},
  primaryClass = {cs.CV},
}

@inproceedings{suvorov2022resolution,
  title={Resolution-robust large mask inpainting with fourier convolutions},
  author={Suvorov, Roman and Logacheva, Elizaveta and Mashikhin, Anton and Remizova, Anastasia and Ashukha, Arsenii and Silvestrov, Aleksei and Kong, Naejin and Goka, Harshith and Park, Kiwoong and Lempitsky, Victor},
  booktitle={Proceedings of the IEEE/CVF winter conference on applications of computer vision},
  pages={2149--2159},
  year={2022}
}

@inproceedings{li2022mat,
  title={Mat: Mask-aware transformer for large hole image inpainting},
  author={Li, Wenbo and Lin, Zhe and Zhou, Kun and Qi, Lu and Wang, Yi and Jia, Jiaya},
  booktitle={Proceedings of the IEEE/CVF conference on computer vision and pattern recognition},
  pages={10758--10768},
  year={2022}
}

@inproceedings{sargsyan2023mi,
  title={Mi-gan: A simple baseline for image inpainting on mobile devices},
  author={Sargsyan, Andranik and Navasardyan, Shant and Xu, Xingqian and Shi, Humphrey},
  booktitle={Proceedings of the IEEE/CVF International Conference on Computer Vision},
  pages={7335--7345},
  year={2023}
}

@inproceedings{jain2023keys,
  title={Keys to better image inpainting: Structure and texture go hand in hand},
  author={Jain, Jitesh and Zhou, Yuqian and Yu, Ning and Shi, Humphrey},
  booktitle={Proceedings of the IEEE/CVF winter conference on applications of computer vision},
  pages={208--217},
  year={2023}
}

@misc{batifol2025flux,
  author       = {Batifol, Stephen and Blattmann, Andreas and Boesel, Frederic and Consul, Saksham and Diagne, Cyril and Dockhorn, Tim and English, Jack and English, Zion and Esser, Patrick and Kulal, Sumith and others},
  title        = {Flux. 1 kontext: Flow matching for in-context image generation and editing in latent space},
  year         = {2025},
  eprint       = {2506.15742}, 
  archivePrefix = {arXiv},
}

@inproceedings{zhuang2024task,
  title={A task is worth one word: Learning with task prompts for high-quality versatile image inpainting},
  author={Zhuang, Junhao and Zeng, Yanhong and Liu, Wenran and Yuan, Chun and Chen, Kai},
  booktitle={European Conference on Computer Vision},
  pages={195--211},
  year={2024},
}

@inproceedings{manukyan2023hd,
  title={Hd-painter: high-resolution and prompt-faithful text-guided image inpainting with diffusion models},
  author={Manukyan, Hayk and Sargsyan, Andranik and Atanyan, Barsegh and Wang, Zhangyang and Navasardyan, Shant and Shi, Humphrey},
  booktitle={International Conference on Learning Representations},
  year={2023}
}

@inproceedings{li2025rorem,
  title={Rorem: Training a robust object remover with human-in-the-loop},
  author={Li, Ruibin and Yang, Tao and Guo, Song and Zhang, Lei},
  booktitle={Proceedings of the Computer Vision and Pattern Recognition Conference},
  pages={14024--14035},
  year={2025}
}

@inproceedings{podell2023sdxl,
  title={Sdxl: Improving latent diffusion models for high-resolution image synthesis},
  author={Podell, Dustin and English, Zion and Lacey, Kyle and Blattmann, Andreas and Dockhorn, Tim and M{\"u}ller, Jonas and Penna, Joe and Rombach, Robin},
  booktitle={International Conference on Learning Representations},
  volume={2024},
  pages={1862--1874},
  year={2024}
}

@inproceedings{rombach2022high,
  title={High-resolution image synthesis with latent diffusion models},
  author={Rombach, Robin and Blattmann, Andreas and Lorenz, Dominik and Esser, Patrick and Ommer, Bj{\"o}rn},
  booktitle={Proceedings of the IEEE/CVF conference on computer vision and pattern recognition},
  pages={10684--10695},
  year={2022}
}

@inproceedings{ma2024imdl,
  title={Imdl-benco: A comprehensive benchmark and codebase for image manipulation detection \& localization},
  author={Ma, Xiaochen and Zhu, Xuekang and Su, Lei and Du, Bo and Jiang, Zhuohang and Tong, Bingkui and Lei, Zeyu and Yang, Xinyu and Pun, Chi-Man and Lv, Jiancheng and others},
  booktitle={Advances in Neural Information Processing Systems},
  volume={37},
  pages={134591--134613},
  year={2024}
}

@article{kadam2021multiple,
  title={Multiple image splicing dataset (MISD): a dataset for multiple splicing},
  author={Kadam, Kalyani Dhananjay and others},
  journal={Data},
  volume={6},
  number={10},
  pages={102},
  year={2021},
}

@article{ci2025diffseg30k,
  title={DiffSeg30k: A Multi-Turn Diffusion Editing Benchmark for Localized AIGC Detection},
  author={Ci, Hai and Peng, Ziheng and Yang, Pei and Xuan, Yingxin and Shou, Mike Zheng},
  journal={arXiv preprint arXiv:2511.19111},
  year={2025}
}

@misc{wu2026locateedit,
  author       = {Wu, Shiyu and Li, Shuyan and Li, Jing and Liu, Jing and Wang, Yequan},
  title        = {LocateEdit-Bench: A Benchmark for Instruction-Based Editing Localization},
  year         = {2026},
  eprint       = {2602.05577},
  archivePrefix = {arXiv},
  primaryClass = {cs.CV},
}

@inproceedings{wang2025opensdi,
  title={Opensdi: Spotting diffusion-generated images in the open world},
  author={Wang, Yabin and Huang, Zhiwu and Hong, Xiaopeng},
  booktitle={Proceedings of the IEEE/CVF Conference on Computer Vision and Pattern Recognition},
  pages={4291--4301},
  year={2025}
}

@article{telea2004image,
  title={An image inpainting technique based on the fast marching method},
  author={Telea, Alexandru},
  journal={Journal of graphics tools},
  volume={9},
  number={1},
  pages={23--34},
  year={2004},
}

\end{document}